\documentclass[11pt]{article}

\usepackage{array}
\usepackage{arxiv}
\usepackage{multirow}
\usepackage{vcell}
\usepackage[utf8]{inputenc} 
\usepackage[T1]{fontenc}    
\usepackage{hyperref}       
\usepackage{url}            
\usepackage{booktabs}       
\usepackage{amsfonts}       
\usepackage{nicefrac}       
\usepackage{microtype}      
\usepackage{graphicx}
\usepackage{natbib}
\usepackage{doi}
\usepackage{float}
\usepackage{subfigure}
\usepackage{algorithmicx,algorithm,float}
\usepackage{amsmath}

\makeatletter
\newenvironment{breakablealgorithm}
  {
   \begin{center}
     \refstepcounter{algorithm}
     \hrule height.8pt depth0pt \kern2pt
     \renewcommand{\caption}[2][\relax]{
       {\raggedright\textbf{\ALG@name~\thealgorithm} ##2\par}%
       \ifx\relax##1\relax 
         \addcontentsline{loa}{algorithm}{\protect\numberline{\thealgorithm}##2}%
       \else 
         \addcontentsline{loa}{algorithm}{\protect\numberline{\thealgorithm}##1}%
       \fi
       \kern2pt\hrule\kern2pt
     }
  }{
     \kern2pt\hrule\relax
   \end{center}
  }
\makeatother

\title{Interpreting Machine Learning Models for Room Temperature Prediction in Non-domestic Buildings}

\author{ {Jianqiao Mao} 
        \thanks{Department of Electronic and Electrical Engineering, Gower Street, London, UK, WC1E 6BT} \\
	University College London \\
	\texttt{jianqiao.mao.20@ucl.ac.uk} \\
	\And
    {Grammenos Ryan $^*$} \\
	University College London\\
	\texttt{r.grammenos@ucl.ac.uk} \\
}

\date{}

\renewcommand{\headeright}{}
\renewcommand{\shorttitle}{\textit{Interpreting Machine Learning Models for Room Temperature Prediction in Non-domestic Buildings}}

\hypersetup{
pdftitle={Interpreting the Machine Learning Model from Data Perspective for Room Temperature Prediction in Non-domestic Building},
pdfsubject={q-bio.NC, q-bio.QM},
pdfauthor={Jianqiao Mao, Ryan Grammenos},
pdfkeywords={Machine Learning Interpretability, Room Temperature Prediction},
}

\begin{document}
\maketitle

\begin{abstract}
An ensuing challenge in Artificial Intelligence (AI) is the perceived difficulty in interpreting sophisticated machine learning models, whose ever-increasing complexity makes it hard for such models to be understood, trusted and thus accepted by human beings. The lack, if not complete absence, of interpretability for these so-called black-box models can lead to serious economic and ethical consequences, thereby hindering the development and deployment of AI in wider fields, particularly in those involving critical and regulatory applications. For example, managing building energy consumption efficiently is paramount to tackling issues of global warming and shortage of energy resources. Yet, the building services industry is a highly-regulated domain requiring transparency and decision-making processes that can be understood and trusted by humans. To this end, the design and implementation of autonomous Heating, Ventilation and Air Conditioning (HVAC) systems for the automatic but concurrently interpretable optimisation of energy efficiency and room thermal comfort is of topical interest. This work therefore presents an interpretable machine learning model aimed at predicting room temperature (RT) in non-domestic buildings, for the purpose of optimising the use of the installed HVAC system. We demonstrate experimentally that the proposed model can accurately forecast room temperatures eight hours ahead in real-time by taking into account historical RT information, as well as additional environmental and time-series features. In this paper, an enhanced feature engineering process is conducted based on the Exploratory Data Analysis (EDA) results. Furthermore, beyond the commonly used Interpretable Machine Learning (IML) techniques, we propose a Permutation Feature-based Frequency Response Analysis (PF-FRA) method for quantifying the contributions of the different predictors in the frequency domain. Based on the generated reason codes, we find that the historical RT feature is the dominant factor that has most impact on the model’s prediction.
\end{abstract}

\keywords{Machine Learning Interpretability \and EDA \and Temperature Prediction \and Building Management \and HVAC}

\section{Introduction}

The exponentially increasing number of Machine Learning (ML) systems recently have been applied to assist humans in daily decision-making, quantitative analysis and complex system controlling. Their promising performance and significant potentials have been observed in fields like healthcare\citep{meira2020contextualized}\citep{char2018implementing}, intelligent business\citep{kraus2020deep}\citep{ebadi2019can}, building management\citep{grammenos2021analysis}\citep{fan2019novel} and to name but a few, attracting increasing attention from both academia and industry. From the traditional machine learning techniques to the state-of-the-art deep learning methods, most of the research focuses on improving the model performance. Benefitting from the continuous and intensive exploration, the predictive capability of the machine learning models is significantly improved, but, at the same time, the model complexity also exponentially increased. For example, the deep learning-based Enhanced Super-resolution Network (EDSR) for single image super-resolution reports its number of parameters is over 43 million with 32 residual blocks\citep{lim2017enhanced}. In fact, it is extremely hard to interpret such a deep network with the complex structure, due to the fact that the more complex a model is, the worse its transparency is expected, which is a fundamental trade-off between the model performance and interpretability\citep{hall2019introduction}. Therefore, an ensuing challenge is that the models are becoming more and more difficult to be understandable for humans, which is the so-called “black-box”.

It seems acceptable to bet more in model performance with a certain loss of interpretability in some less critical applications such as interactive video games, entertainment services and language translations. However, as machine learning-based systems playing an increasingly important role in more critical areas, the need of understanding and trusting the model itself and modelling results has become essential\citep{mi2020review}. On the one hand, some regulatory industries like medicine, energy and finance must understand why and how a machine made a decision that can significantly affect our lives or even the human society and history\citep{hall2019introduction}, and thus doctors, engineers and administrators can refer to the modelling results with certain trust. Furthermore, without Interpretable Machine Learning (IML) techniques to open the complex black boxes for a deep insight into their prediction logics, there would be serious consequences. For example, a rising concern is if the black-box model is fragile to potential adversarial attacks. Previous studies have shown that in autonomous driving, slight changes (even for a few pixels) in road signs’ image can lead to completely different recognition results\citep{deng2020analysis}, which might be a fatal flaw potentially leading to serious safety and ethical consequences. Meanwhile, interpretability is also a guarantee to enable the applied machine learning model to unbiasedly operate in scenarios of recruitment, college admission, criminal conviction, to name but a few\citep{hall2019introduction}. In many cases, although there is no original intention for engineers to introduce human bias, some unforeseen and regrettable bias is re-learned by the model, due to the originally biased data or the inappropriate way to select features.

More specifically, this paper attempts to investigate the validity of applying machine learning techniques in the building service industry with more emphasis to deal with the aforementioned challenges of interpretability and model bias. It has become more and more critical to carefully manage energy consumption and improve energy efficiency, because they highly correlate to many global challenges such as global warming and energy shortage. According to the International Energy Agency (IEA), the final energy usage in buildings soared from 118EJ in 2010 to 128EJ in 2019, responsible for 28\% of the global energy-related carbon dioxide ($CO_2$) emissions\citep{trackBuilding2020}. In particular, the Heating, Ventilation and Air Conditioning (HVAC) systems are a key point to be investigated, because they not only consume more than a half building energy\citep{chua2013achieving}, but also can significantly influence the productivity of individuals through controlling the thermal comfort levels\citep{shaikh2014review}. Therefore, many researchers attempted to not only optimise the energy efficiency but also intelligently tune the system to reach the thermal comfort by applying various powerful machine learning techniques. For example, researchers in\citep{satrio2019optimization} proposed a hybrid method using Artificial Neural Network (ANN) and multi-objective genetic algorithm, which helps the design and selection of the control strategy that can operate the HVAC systems efficiently. Comparably, Andrew et al. ensembles four multiple-linear perception (MLP) models, reducing the energy consumed by HVAC system by more than 7\%\citep{liu2019novel}. In our previous work\citep{grammenos2021analysis}, a self-adaptive HVAC system design is presented to optimise system energy efficiency with respect to the occupants’ setpoint preference. 

Although a number of machine learning techniques are used to deal with this problem, the absence of model interpretability also prevents further success in the wider applications. Importantly, a comprehensive understanding for a given machine learning model is able to inform further model optimisation and figure out the reason why a model correctly or incorrectly predicts the target. To efficiently optimise a machine learning model, feature engineering is one of the most critical steps to ensure models accurate, generalisable and computationally efficient. However, in the big data era, more diversifying features may have potential contributions, which, at the same time, makes the feature selection process more demanding. It is indeed that introducing a greater number of features usually boosts (at least not weaken) the model performance, but some of the irrelevant features can not only potentially poison the model and generate a spurious association relation but also make the model cumbersome and computationally costly. Without proper interpretability techniques to look deeper into how the model behaves and responds to certain features, the potential improvement by applying the currently more and more complicated modelling methods would be challenging and even limited. In addition, business owners often cannot bear the economic and security risks of applying unknown black-box models to such a critical area of HVAC systems management. For example, when the Room Temperature (RT) predicted by an autonomous HVAC system based on a black-box machine learning model conflicts with the system administrator’s predictions based on his/her experience, the reason of why the model generates its prediction is always wanted to justify if it is reasonable. In other words, they need interpretable machine learning techniques to help them understand the process of how the models make a decision and thus build the confidence to trust these important business decisions that may affect both the workers productivity and building energy costs.

In this paper, we aim to explore various state-of-the-art IML techniques to inform how an effective and trustworthy machine learning model can be designed and applied to reasonably manage HVAC systems. Using a dataset collected from a non-domestic office building including observations from outdoor and indoor environmental sensors and HVAC system states indicators, a deeper exploration of the data is conducted through various Exploratory Data Analysis (EDA) techniques firstly, revealing intrinsic characteristics of the data to inform the afterwards feature engineering and model optimisation. Trained on the given dataset, it is proved possible to accurately predict the RT thus indirectly informing how to energy-efficiently control the HVAC system. With the support of Interpretable Machine Learning (IML) techniques, the model global and local behaviours are analysed to validate its performance and robustness and reveal potential risks when predictions deviate from the ground true. Briefly, the contributions of this work are summarised as follow:

\begin{itemize}
	\item \textbf{i) The understanding of the dataset is enhanced by applying a set of EDA techniques, which guides the feature engineering process.} By applying the EDA on Indoor, Outdoor and Time Series (IOTS) features and the target variable, their characteristics such as distribution, stationarity, dependency and structural change are comprehensively explored and analysed. The results inform the subsequent feature engineering process.
	\item \textbf{ii) The enhanced feature engineering significantly improves the prediction accuracy.} By introducing new features such as the historical moving average of the target variable and the holiday indicator, the fine-tuned Extreme Gradient Boosting Machine (XGBM) regressor achieves Mean Absolute Error (MAE) of 0.9263°C equivalent to Mean Absolute Percentage Error of 4.01\% on the test set to predict 8-hour-ahead RTs. Comparing to the one trained with original IOTS features, its predicting error on the validation set is reduced by around 41\%.
	\item \textbf{iii) Except for the commonly used IML techniques, the \textbf{Permutation Feature-based Frequency Response Analysis (PF-FRA)} is proposed to enhance the model’s global interpretability.} Although the existing IML techniques are able to provide abundant explanations both globally and locally to open the black box, most of them, to our best knowledge, only focus on their time-series properties. The proposed \textbf{PF-FRA} aims to quantify the feature contributions in frequency domain, which can help users understand which feature mainly contributes to the high-frequency and Direct Current (DC) components of the model response, leading to fluctuations or magnitude.
\end{itemize}

The rest of this paper is organized in the structure shown as follow. Section 2 widely reviews and summaries a variety of related research of applying machine learning techniques in predicting non-domestic buildings’ RT. In Section 3, after briefly introducing the used dataset in this paper, we present the enhanced feature engineering motivated by the results of exploratory data analysis, and how to select and interpret the desired machine learning model. Section 4 demonstrates the process to finalise the XGBM regressor and explains the model with the support of existing and proposed IML techniques. In Section 5, the significance of this research is concluded and potential future work is then indicated.

\section{Literature Review}

Indoor RT prediction is expected to help HVAC systems to optimise the energy efficiency with the respect to the thermal comfort in the non-domestic buildings by providing important information for advance control. Many investigations recently focus to use machine learning techniques, including traditional statistical learning and more advanced deep learning, to set up a robust and accurate predicting model. This section reviews some of the related research and discusses their advantages and limitations.

Traditional machine learning models are proved able to predict the indoor RT adaptively and accurately. Paul et al. in their study \citep{paul2018iot} compares the indoor temperature predicting performance among three of the most popular predictive regression models: Random Forest (RF), Support Vector Machine (SVM) and Neural Networks (NN). Trained with an online learning strategy, these models generate their prediction based on a number of indoor features such as $CO_2$ level, lighting strength and humidity and outdoor features including outdoor air temperature, solar irradiance, wind velocity, facade illuminance and to name but a few, captured by real sensors. Although the NN achieves lower mean bias error at 0.6565 compared to the figures of the others, the RF model has the lowest coefficient of variance at 0.0840 with the best R-square score at 0.9855. Based on the models’ promising predictive ability, the selected machine learning model is integrated into an edge computing-based IoT system to optimise the energy efficiency for a building. Comparably, Arendt et al. investigates a variety of machine learning models with different levels of interpretability, categorized as so-called white-box, gray-box and black-box models for a university building\citep{arendt2018comparative}. With the loss of model transparency, the most accurate black-box model in the option pool outperforms the best gray- and white-box models by 0.3°C and 0.6°C, respectively. However, the depending dataset of this research only covers a month period, which provides limited reliability in terms of model robustness. As a more comprehensive study comparison among various learning models\citep{alawadi2020comparison},  Alawadi et al. compares 36 models from 20 different algorithm families, including Bayesian, ensembles, linear, Gaussian, NNs and to name but a few. A similar experimental results like research \citep{paul2018iot} indicating that the tree-based regressors show a better performance metrics of R-square score than NN-based models and SVM. Researchers clarify that the possible reason is the tree-based model tends to be less noise-sensitive, and thus it is likely to have a better robustness for outliers than the more complex models like NN. However, their experiments do not involve any features or indicators from the HVAC system, which means its validity and generalisation need confirmed by further investigations.

On the other hand, thanks to the continuous explorations for the NNs’ design and optimisation, many promising variants are successfully implemented to predict the indoor air temperature in some recent research. In research \citep{li2013hvac}, a Back Propagation NN (BPNN) is used by Li et al. to predict the temperature of a room with an operating HVAC system. The study argues the challenges of traditional modelling methods to perform an accurate control due to the non-linearity of the HVAC system, while a BPNN is designed to approximate the non-linear system with promising performance. The implemented NN-based predictive control strategy is experimentally validated in this research, which is able to forecast and control the RT with the respect to energy efficiency. In an earlier study \citep{ruano2006prediction}, a simple NN is used to perform long-term indoor temperature prediction, while  multi-objective genetic algorithms are applied to optimise the model parameters instead of gradient descent used in \citep{li2013hvac}. The proposed RT predicting model are believed to be extendable with satisfactory predictions on a one-year period with the help of an adaptive sliding window. However, it is regret that simulations only cover a half month in winter, which needs further validations for the model robustness with longer range of data. A potential problem of the aforementioned investigations is that they do not consider the indoor temperature as a time series, while abundant evidence has shown that treating the temperature as a time series is supposed more reasonable \citep{mateo2013machine} \citep{godahewa2020simulation} \citep{xu2019improving}.

Therefore, a set of recent research attempts to predict the RT by using time-series analysis methods. Mateo et al. in their investigation \citep{mateo2013machine} proposed a hybrid model combining a Multilayer Perceptron (MLP) with a Non-linear Autoregressive Exogenous (NARX) for indoor air temperature prediction. Evaluated by MAE, it achieves the best predicting performance comparing to not only the traditional machine learning models such as Robust Multiple Linear Regression (RMLR), Extreme Learning Machine (ELM) and Autoregressive Exogenous (ARX) but also a simple MLP model. Noticeably, they further introduce a clustering model to group the observations with higher similarities, and thus an expert model is trained and applied for each cluster group, which is expected to enhance the models’ specificity. As a more advanced time series analysis model, Recurrent Neural Networks (RNN) suits the RT prediction task with promising predictive capability. In the research \citep{godahewa2020simulation}, a RNN is introduced to predict the temperature of a university’s lecture theatre when it is occupied. It experimentally outperforms other machine learning models like SVM, RF, MLP and Feed-Forward Neural Networks (FFNN) in terms of Root MSE (RMSE) for its predictions. Simulations on the real data in this work further demonstrate that implementing the optimised RNN model in the HVAC controlling system contributes to around 20\% more of energy saving according to its accurately forecasting, compared to traditional system controlling techniques. However, one of the recognised drawbacks of RNN is that it can only consider it previous outputs within a short term, and thus it tends to fail in predicting a time series depending on relatively long historical patterns. As a variant of RNN, Long-Short-Term Memory (LSTM) model is designed to overcome RNN’s limitations. A recent work \citep{xu2019improving} conducted by Xu et al. presents a novel modified LSTM model for 5-min and 30-min ahead indoor temperature prediction, which slightly improves the predicting performance compared to other traditional machine learning and other NN-based models. Similar to some of the aforementioned research, since only observations of one week in summer are tested in this case, limited period for testing needs further validation in terms of model robustness for different seasons and years.

\begin{table}[htbp]
  \centering
  \caption{Summary of Reviewed Literature}
  \begin{tabular}{c}
    \begin{minipage}[b]{1\columnwidth}
		\centering
		\raisebox{-0.5\height}{\includegraphics[width=\linewidth]{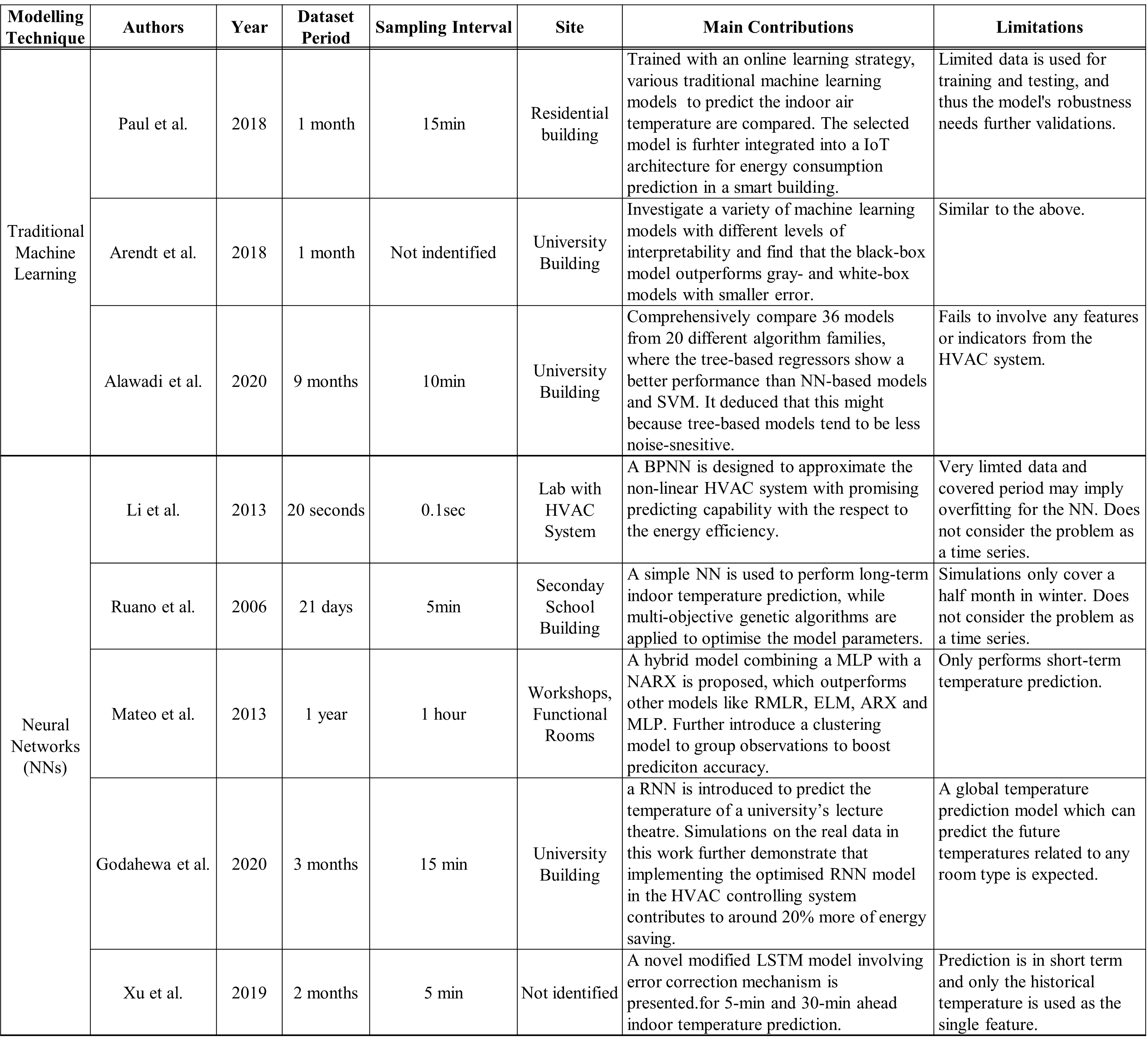}}
	\end{minipage}
  \end{tabular}
\label{literatureSummary}
\end{table}

Table \ref{literatureSummary} summarises the reviewed related work that attempts to apply machine learning techniques to deal with the task of indoor temperature prediction by traditional machine learning and NN-based models. It is worthy to recognise that despite witnessed significant improvements using more and more complex black-box models such as SVM and NNs, loss of interpretability and trust can lead to safety and ethical risks discussed in the previous chapter. In particular, the reviewed data-driven approaches learn the associational relations between features and the RT only depending on observational data, which generates unreliable predictions. Therefore, we are also interested in opening these black boxes by certain IML techniques to explain the model mechanism thus enhancing trust and model reliability.

\section{Methodology}

    \subsection{Dataset Description and Analysis}
    
    The original dataset explored in this paper contains 3-year observations from both indoor and outdoor sensors and system indicators of an HVAC system installed in an 11-floor non-domestic office building located in Athens, Greece, provided by \textit{General Technology Ltd.} The dataset records observations in the period from December 8th, 2017 to September 18th, 2020. For the indoor sensors and HVAC system indicators, the Setpoint Temperature (SPT), the HVAC system On/Off state, the system operation mode state and the building alarm signal state are sampled on a 10-min basis, while for the outdoor sensors, the outdoor temperature and humidity are measured on a state-change basis. Importantly, the target variable RT is recorded in the granularity of 0.5 degree instead of continuous values due to the limitation of the temperature sensor’s sensitivity.

    \begin{table}[htbp]
      \centering
      \caption{Summary of Features in the Pre-processed Dataset}
      \begin{tabular}{c}
        \begin{minipage}[b]{1\columnwidth}
    		\centering
    		\raisebox{-0.5\height}{\includegraphics[width=\linewidth]{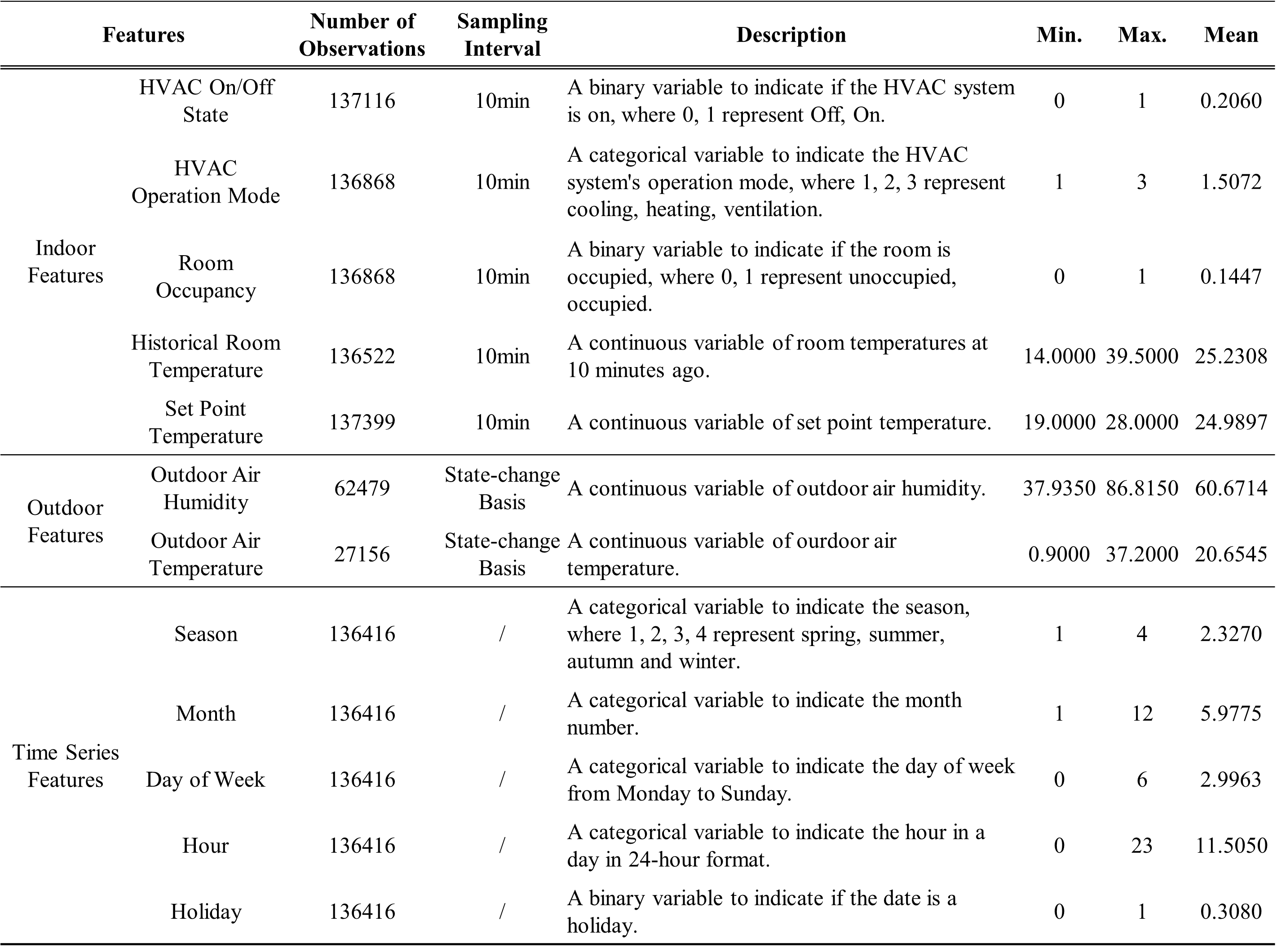}}
    	\end{minipage}
      \end{tabular}
    \label{DatasetSummary}
    \end{table}
    
    Based on the automatic data cleaning and preparation techniques proposed in our previous work, the original dataset is processed to a clean and structural dataset with aligned timestamps and extracted features. Noticeably, an additional feature named Occupancy State is created accordingly to indicate if the room is possibly occupied and some time series features representing samples’ time-series characteristics are added. Moreover, the alarm signal state in the original dataset is excluded due to uncorrelation to the predicting the targeted RT. Considering the special circumstance of the COVID-19 pandemic breakout, the observations after February 29th, 2020 are excluded from this paper to maintain the consistency of the data pattern. Splitting the overall dataset on two dates: June 30th, 2019 and October 10th, 2019 divide the dataset into training set covering the period from December 8th, 2017 to June 30th, 2019, validation set covering the period from July 1st, 2019 to October 10th, 2019 and test set covering the period from October 11th, 2019 to February 29th, 2020. The information of features in the pre-processed dataset is summarised by Table\ref{DatasetSummary} below, while it is worthy to note that the holiday indicator and the historical room temperature feature are newly introduced and the previously included year feature is excluded in this work.
    
\begin{figure}[htbp]
    	\centering
    	\includegraphics[width=1\textwidth]{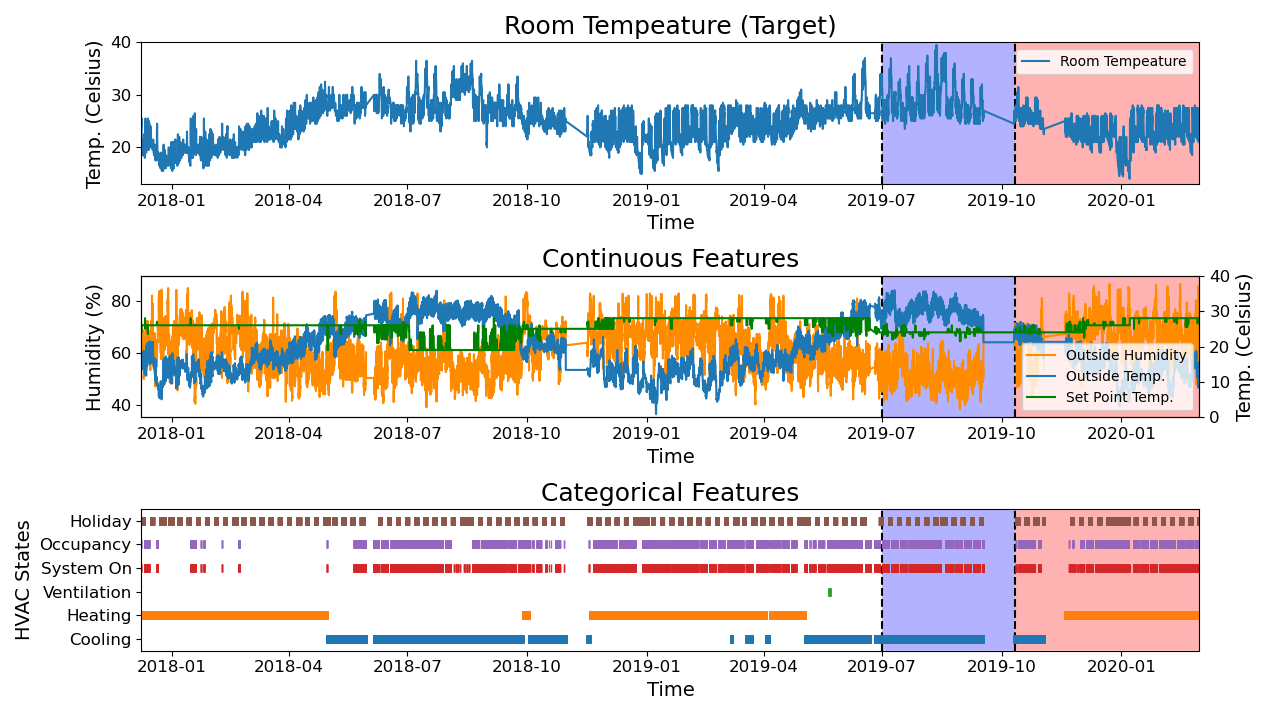}
    	\caption{Visualisation of Data and Splitting}
    	\label{DataVisualisation}
    \end{figure}

    
    \paragraph{Data Visualisation}
    Data visualisations help to intuitively perceive the data properties, for example, time-series trends, correlating relationships and fluctuating ranges. Fig.\ref{DataVisualisation} illustrates features in the processed dataset including the target RT (the 1st subplot), continuous features (the 2nd subplot) as well as categorical features (the 3rd subplot) in time-series representations, where the periods filled with blue and red are divided as validation and test sets and the remaining part is used to train the machine learning model.
    
    It can be observed that the target variable RT has obvious seasonality ranging from the lowest 14.0 degrees to the highest 39.5 degrees, that is, its values tend to fluctuate around a lower mean value in winter but around a higher mean value in summer. Noticeably, the RT periodically fluctuates day by day like features of outdoor air temperature and humidity, and its trend is similar to the former but with less violent changes. On the contrary, the SPT remains during many time periods with occasional changes but an opposite trend to the RT. These phenomena match our experience that the indoor RT is expected higher when the outdoor temperature is also relatively higher, and room occupiers tend to set a lower SPT when the RT is measured higher and verse versa. Comparably, the outdoor humidity shows an inversed trend to the outdoor air temperature, which might because of the fact that humidity and temperature in the natural environment are always negatively correlated. As for the categorical features, similarities are found between the created occupancy state and the system On/Off state, while the holiday feature is periodical with certain weekly and yearly repetitions. Besides, the remaining three indicators indicate the HVAC system’s operation mode in one of the ventilation, heating and cooling states.
    
    \paragraph{Data Imbalance}
    As discussed, visualisations have shown the target variable RT has certain short- and long-term seasonality, but its deeper patterns still need more advanced explorations. Obviously, some high extreme values especially in summer and low extreme values especially in winter which fluctuate around its trend line can be found. We care about the distributions of the target RT to check if there is sample imbalance, because it can affect the afterwards modelling technique selection.
    
\begin{figure}[htbp]
    	\centering
    	\includegraphics[width=1\textwidth]{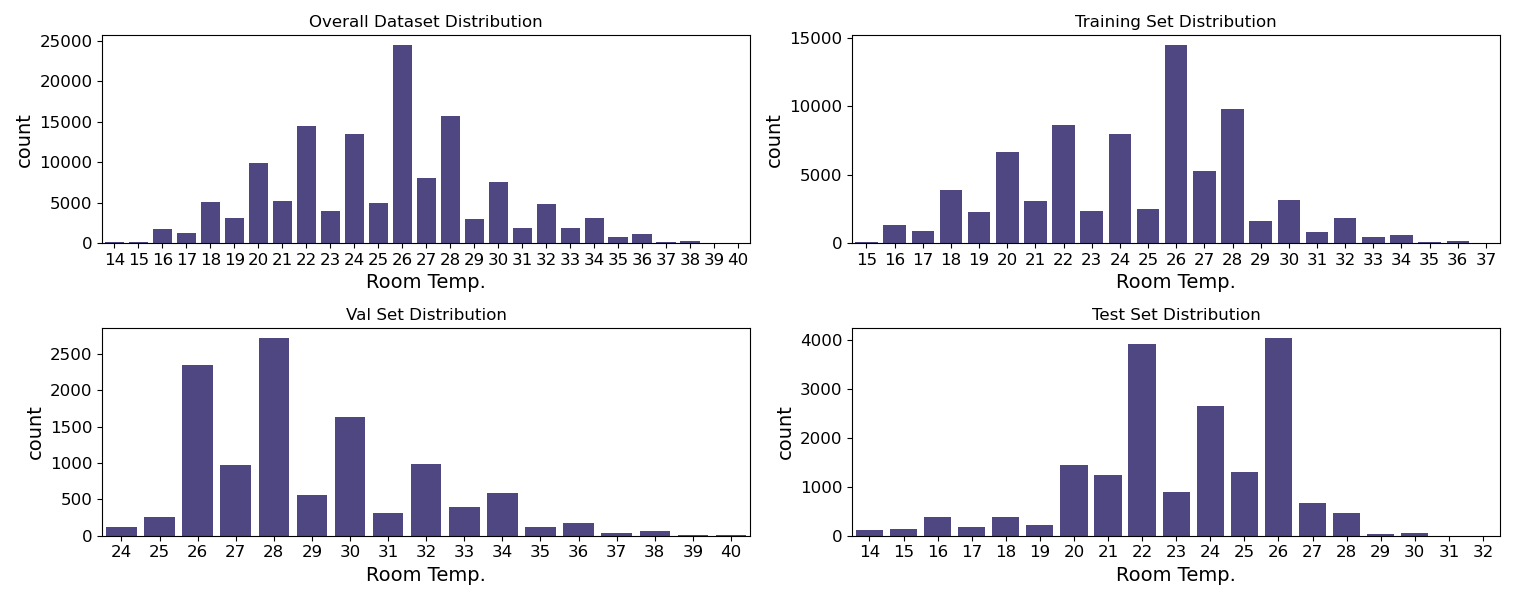}
    	\caption{Target RT Distributions on Different Sets}
    	\label{TargetDistribution}
    \end{figure}

    Most machine learning models expect not only the input features but also the target variable to obey normal distribution, so the classes of observations always need to be balanced. Exploring the target distribution before modelling should inform if the data balancing techniques are needed to prevent the constructed model from bias. Fig.\ref{TargetDistribution} demonstrates distributions of the target RT values with the granularity of 1 degree Celsius for the overall dataset and the split sets.
    
    Obviously, the distributions for different sets are significantly imbalanced and even different from each other. On the one hand, although the training set follows the most similar distribution of the overall dataset, the target distributions of validation and test sets exclusively differ. For the overall dataset and training set, the count of 26 degrees is the highest and the 28 takes the second place, while 30 and 32 degrees take accounts a big proportion in the validation set and 22 and 24 degrees are the 2nd and 3rd most frequent values. The reason caused the significant difference in target distributions for these sets is the covered periods. For example, the training set contains 1.5 years of observations which cover all the seasons, but the covered periods for validation and test sets are only around a quarter (roughly autumn and winter, respectively). On the other hand, all of these sets suffer from severe class imbalance, which means a few classes account for the majority of observations. This imbalance is likely to limit the machine learning model performance because the model tends to treat majority classes with higher degree of importance. Therefore, efforts are needed to mitigate the negatives due to the imbalanced distributions in the subsequent stages.
    
    \paragraph{Target-Features Dependency}
    Most of the current machine learning techniques attempt to learn the associated relationships between the input features and the output target, and thus it is helpful to preview the relations between input features and the target before building machine learning models. For simplification, we only investigate the dependency between the continuous features and the target RT, since there are only limited values for categorical and binary features in our dataset even much less than the number of target’s possible values. With three combinations of any two of the continuous features, the joint relations to the target RT are demonstrated by plots in Fig.\ref{TargetFeature} below. Note that the joints relations are discussed based on the observations in different sets that are divided accordingly to intuitively demonstrate if there is a change of the joint relation between features and target.
    
    \begin{figure}[htbp]
    	\centering
    	\includegraphics[width=1\textwidth]{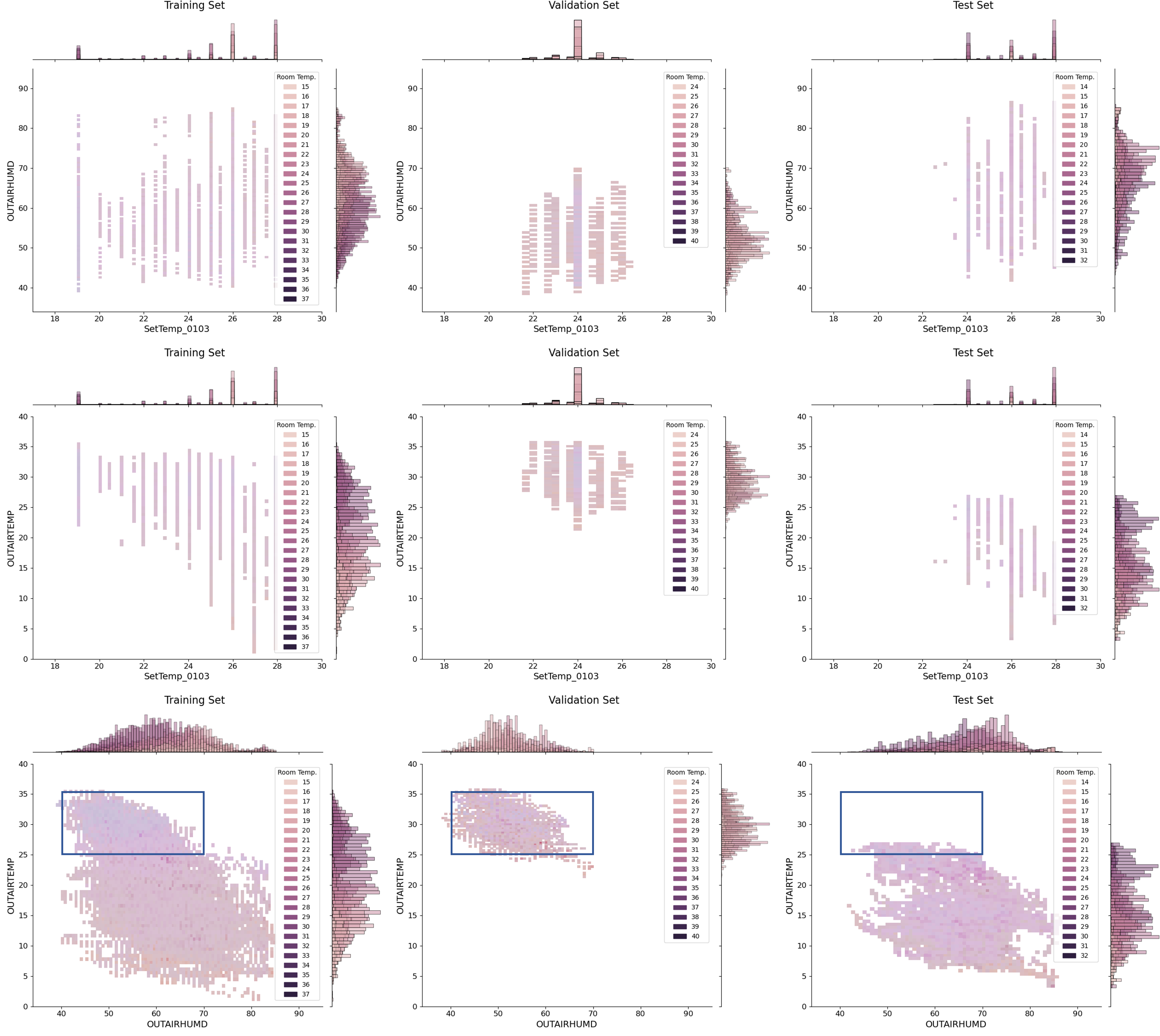}
        \caption{Joint Relation Plots for Continuous Features and Target Room Temperature}
        \label{TargetFeature}
    \end{figure}
    
    The joint dependency plots in Fig.\ref{TargetFeature} indeed show certain relations between the target RT (simplified in 1 degree granularity) and the continuous features in the dataset, but the relations do not seem to be consistent across the different divided datasets. In other words, significant differences can be observed for not only the target and feature distributions cross values but also the corresponding associate responses of the target RT to features in the split sets of training, validation and test. For example, the third row of Fig. 6 illustrates the dependency between the two outdoor features and the SPT values. Comparing the same range of outdoor temperature from 25 to 35 degrees and of outdoor humidity from 40\% to 70\% for training and validation sets outlined by the blue boxes, the majority of RT values in the training set in this area range from 26 to 29 degrees, while they range from 29 to 32 degrees in the validation set. Moreover, there is only few observation in this outlined area for the test set.
    
    \subsection{Enhanced Feature Engineering}
    
    Our previous work has demonstrated advancement in feature engineering by introducing additional time-series and room occupancy state features and removing the uncorrelated alarm signal feature, which indeed boosts the model performance. Therefore, in this work, these features are reproduced in the same way. Meanwhile, setting the RT as the target variable, the SPT should be excluded from the feature set in the following modelling process due to the aim that we want a predictive model that can inform the SPT adjustment according to the prediction of RT.
    
    In this work, it is not possible to acquire more features from other sensors and system indicators with the given dataset collected in the aforementioned period, so more efforts focus on extract useful features based on the existing ones. Observing violent fluctuations for the continuous features which may enable the model to learn patterns from these high frequency noises, the moving average filter is applied to smoothen the outdoor temperature and humidity and a newly introduced historical RT feature (define and discuss later). Furthermore, based on the previous work’s feature engineering on Indoor, Outdoor and Time Series (IOTS) features, enhanced modifications are made to improve the model performance.
    
    \paragraph{Moving Average Filter}
    Previous analysis has shown that violent fluctuations are observed in all continuous features and the predicting target RT. Although fluctuations seem to provide more abundant information for a given feature, they also can lead to the overfitting problem for time series analysis. Because the short-term violent fluctuations can be too noisy and the data-driven machine learning model usually tends to learn the pattern from noise in this case, instead of from the more important mid-term and long-term trend information, these high-frequency components usually negatively affect the robustness of the model. A typical and efficient approach to eliminate the effects of high-frequency noise is Moving Average (MVA) filter\citep{box2015time}.
    
    \paragraph{Indoor, Outdoor and Time Series (IOTS) Features}
    The Indoor, Outdoor and Time Series (IOTS) features in the given dataset contain indoor SPT and HVAC system indicators, outdoor humidity and temperature, time-series features like year, month, day and hour. Briefly, we modified the features by removing, adding and filtering. Specifically, the SPT and year features are removed from the dataset, the feature of holiday indicator is introduced, and continuous features are smoothened by MVA filters.
    
\begin{figure}[htbp]
    	\centering
    	\includegraphics[width=1\textwidth]{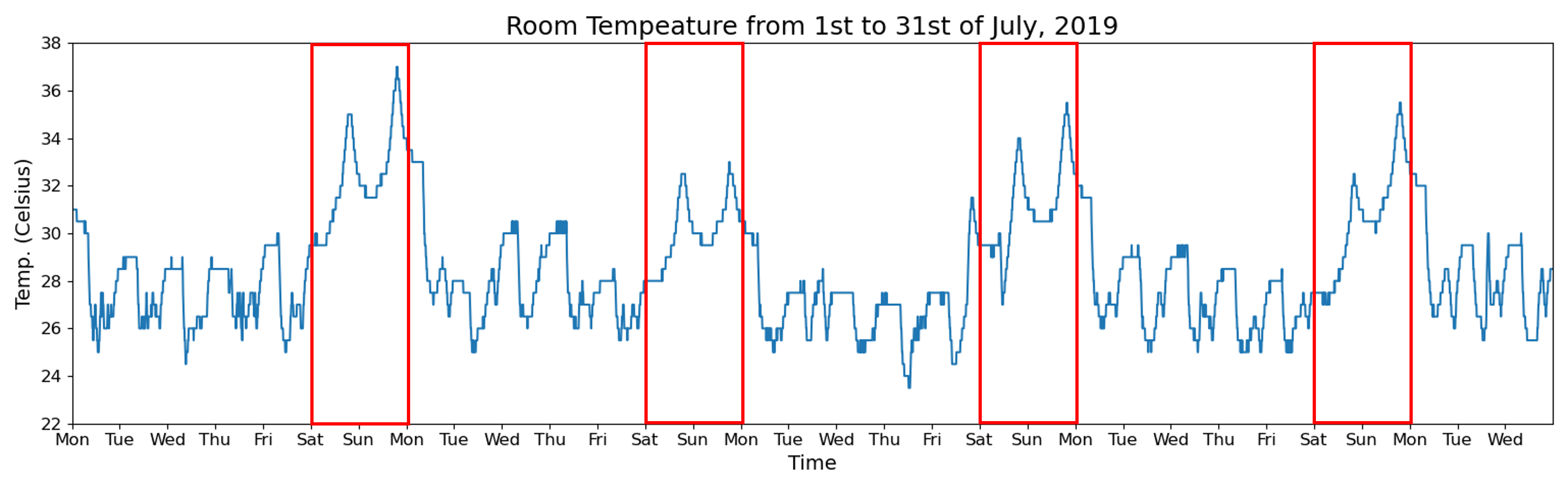}
    	\caption{Room Temperature in July 2019}
    	\label{WeeklySeasonality}
    \end{figure}

    For the feature removal, firstly, since this work attempts to build a model that can predict the RT to inform the SPT adjustment with the respect to the energy efficiency and thermal comfort, the SPT feature should be removed from the dataset. Secondly, considering that the covered period of the given dataset is only 3 years, the time-series feature year fails to provide sufficient variability. Meanwhile, in such a short-term period, we cannot see any significant yearly trend in the visualisation step shown in the previous section, and thus it is also excluded from the feature set. Furthermore, the holiday indicator is added as a new feature to indicate if the date is a public holiday (including both the weekends and the national holidays in Greece), since it is found that there are significant holiday effects for the RT’s value, that is, the RT tends to be higher during holidays but lower during working days. For example, Fig.\ref{WeeklySeasonality} investigates monthly observations in July 2019, where the RT reaches a higher peak in the weekends also with higher mean values comparing to the other days. Besides, the MVA filter with the window width of 1 hour, equivalent to 6 observations given the sampling interval of 10min, is applied for the outdoor humidity and temperature to compress noise and emphasise their trends instead of extreme value.
    
    Noticeably, one of the fundamental assumptions for time series predicting is the historical data have the predictive capability to forecast its future trends, which means the predicted time series variable should maintain its intrinsic characteristics over periods. Otherwise, the predicting model is expected to fail in generating reliable forecasts due to the change of properties of the time series itself, so-called stationarity. Although it is intuitive to judge that the target variable RT is not stationary by directly observing Fig. 2, we need a quantitative analysis to confirm this fact, for example, the autocorrelation and partial autocorrelation coefficients and Augmented Dickey-Fuller (ADF) unit root test\citep{dickey1981likelihood}.
    
\begin{figure}[htbp]
        \centering
        \subfigure[Target RT Series]{
        \begin{minipage}[b]{0.45\textwidth}
        \includegraphics[width=1\textwidth]{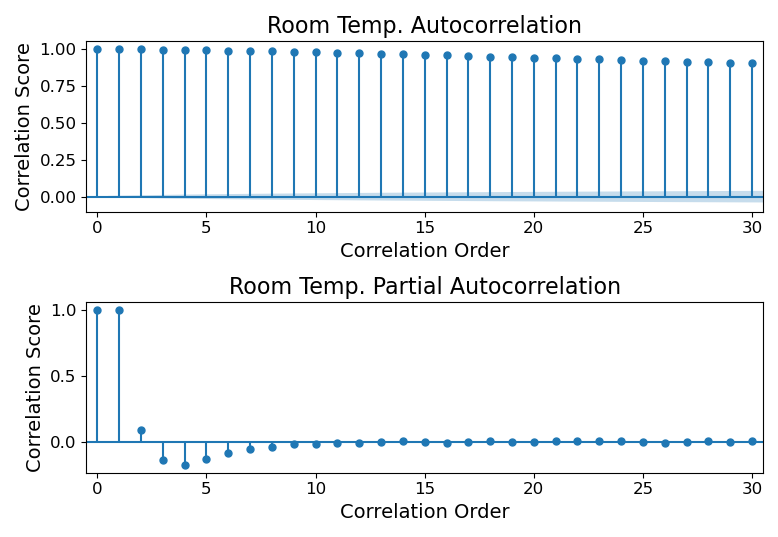} 
        \end{minipage}
        }
        \subfigure[RT’s 1st Order Difference]{
        \begin{minipage}[b]{0.45\textwidth}
        \includegraphics[width=1\textwidth]{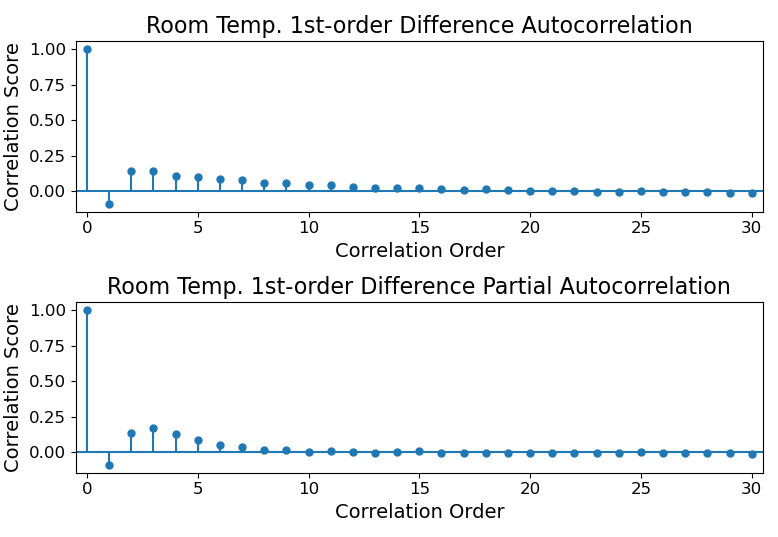} 
        \end{minipage}
        }
        \caption{Autocorrelation and Partial Autocorrelation Plots}
        \label{ACC_PACC_Plot}
    \end{figure}

    On one hand, the autocorrelation and partial autocorrelation of the target RT variable is studied. The AutoCorrelation Coefficients (ACC) with different orders indicate the dependency of a RT value to all of its past values before the given time points, while the Partial AutoCorrelation Coefficients (PACC) with different orders aim to compute the dependency of an observed value to its previous value at a certain time point. Fig.\ref{ACC_PACC_Plot} a) below shows the calculated ACC and PACC of the RT series with orders ranging from 0 to 30. Only a slow decay for the RT’s autocorrelation coefficients as the order increasing is observed, which may imply the non-stationarity. Noticeably, the partial autocorrelation coefficients dramatically converge to roughly 0 after the 1st order, so the RT is deduced to be a random-walk series whose first order difference should be stationary. For further confirmation, we also check the ACC and PACC of RT’s 1st order difference shown in Fig.\ref{ACC_PACC_Plot} b), where both of them converge fast to around 0 at and after the 1st order. That matches our deduction that the RT itself a random walk series because itself is non-stationary but its first order difference is, instead, stationary.
    
    Statistically, the ADF unit root test is applied as a hypothesis test that can indicate if a time series is stationary by setting the null hypothesis \textit{$H_0$: There is a unit root}. and the alternative hypothesis \textit{$H_1$: There is no unit root}. To statistically build the confidence to declare that the RT is a random walk, the ADF unit root test is applied in this case for both the RT series and its 1st order difference. The ADF test result shows p-values for the two tests are $0.12$ and $1.6\times10^{17}$, respectively. Therefore, the null hypothesis \textit{$H_0$} cannot be rejected for RT’s test but should be rejected for RT’s first order difference’s test, which means the RT series is non-stationary but its 1st order difference is stationary. In other words, the hypothesis test results further prove the RT series is a random walk series.

    Based on the stationarity analysis for the target RT variable, we have argued that the RT series is likely to be a random walk series, which means its value at a certain time point is highly correlated to its previous value (at least the most recent one). This kind of time-series property shows a high similarity for the stock price in the financial market. For example, research\citep{agwuegbo2010random} has proved that the representative stock indexes in not only the developing countries such as India but also the developed countries such as the United States and the United Kingdom obey a random walk series which is characterised by a slow decay of the autocorrelation coefficients as the order increasing and only a high 1st order partial autocorrelation. Typically, models to predict stock prices in the financial industry consider the historical information of the price itself as an important input feature\cite{ariyo2014stock}\cite{jain2016dynamic}. Inspired by this, thus it is deduced that the historical RT should also significantly contribute to the future RT prediction.
    
    In this work, instead of introducing individual historical RT values, the MVA filter with the window width of 1 hour is applied in this case to smoothen the original RT series. However, since our predicting target is the RT, it would be a paradox if the historical RT’s moving averages are continuously accessible. Therefore, although the true moving averaged historical RTs are kept on the training set to train the model to approximate to the true data pattern, this feature on the validation and test sets is prepared in a way that moving average value is calculated on a rolling basis according to the model’s self-predicted RTs.

    \subsection{Machine Learning Model}

    In the previous work, a Majority Voting Classifier is applied to predict the SPT because of the poor granularity and thus the limited number of classes. However, we argue that classification models may still not be a suitable approach to predict the RT, since  the classification models tend to treat the target variable as a categorical type without orders and ranks, which causes the loss of the temperature’s relative magnitude information. In fact, the relative magnitude information is important to be learnt by the predicting model and can provide a fairer evaluation of the results. Therefore, instead of using the classifier, the regressor is applied due to its aim to minimising the predicting error in the mean.
    
    As discussed in previous sections, there is much successful related work that applies more advanced deep learning-based methods for indoor temperature prediction more recently with accurate and satisfying prediction. However, the loss of interpretability also brings risks, which make the deep-learning model unexplainable, un-trustable and unreliable. On the other hand, although the white-box models like linear regression and decision tree are transparent and easy to understand their decision-making logic, their predictive capability is very limited for non-linear and high-dimensional problem. Therefore, a tree-based additive boosting model, Extreme Gradient Boosting Machine (XGBM) proposed in \citep{chen2015xgboost}, is applied and investigated in this paper, so the machine learning interpretation discussed in this paper specifies to explain the trained XGBM model.
    
    \subsection{Interpretability Techniques}
    
    The aforementioned EDA, feature engineering and machine learning modelling techniques are expected to help us obtain a satisfactory predictive model that can be applied in the HVAC systems to predict the RT accurately and reliably, but challenges of unknowing how the black-box XGBM regressor generates predictions still remain. That means we are still unable to explain the model behaviour accordingly to improve its transparency and thus enhance human’s trust in the predictions made by the “black box”.
    
    To make the XGBM model somehow explainable, IML techniques should be used to help us understand the model mechanism of decision making, and thus build the confidence to trust the model and its decision. Our discussions mainly focus on two folds of mainstream IML techniques, that is, global and local interpretability. Generally, the global techniques attempt to reveal the internal logic of the black-box models to increase their transparency and the local techniques fit well to finding the causal relationships between the independent and dependent variables for a given data sample\citep{du2019techniques}. To be specific, after we significantly improved the predicting performance by optimising the feature engineering process and carefully selecting a proper machine learning model, the efforts to improve the model interpretability should focus on the reason why the baseline model fails in producing accurate predictions and how the newly introduced features aid it from its poor performance.
    
    Therefore, for the global interpretability, the linear regression- and tree-based surrogate models are firstly applied to approximate the black-box model to reflect its internal mechanism when making predictions, and then Partial Dependency Plots (PDPs)\citep{friedman2001greedy} is applied to analyse how the model’s prediction varies when a certain value of the investigated feature is set. Importantly, Except for the commonly used global IML techniques, we apply the proposed \textbf{PF-FRA} to analyse the frequency components contributed by the IOTS features as well as the newly introduced historical RT feature. Different from most of the existing IML techniques which tend to explain the model under the time-series consideration, the proposed \textbf{PF-FRA} attempts to find out how the interested feature improves the model accuracy in frequency domain.
    
    Although the presented feature engineering and machine learning modelling techniques are expected to have the promising performance to generate as accurate as possible predictions, model bias may still exist especially for those suddenly fluctuating and extreme observations which deviate from the trend line of RT. Analysing the reasons why the built model fails in predicting these observations and comparing them with those accurately predicted observations can be very helpful and informative to further optimise the model’s predictive capability. Therefore, local explanations should be generated for these individual observations to figure out not only the model’s local behaviours but also the logic of how the model predicts some of the RTs inaccurately. For comparison, we select a set of accurate and inaccurate predictions with the same true RT values, where the accurate prediction means the predicted RT value has a less than 0.01-degree difference from the true value, but the inaccurate prediction is defined as the predicted value that deviates from its true value by larger than 2 degrees. In this work, Local Interpretable Model-Agnostic Explanations (LIME)\citep{ribeiro2016should} is applied to analyse the local model behaviours around the selected accurately and inaccurately predicted observations and Shapley Additive Explanations (SHAP)\citep{lundberg2017unified} is used to quantify the importance of features that lead to the unwanted bias.
    
    \paragraph{Global Surrogate Models}
    To understand a black-box model’s logics of decision making, one of the best approaches is to build a simple and interpretable surrogate model to approximate the complex model behaviour and explain the surrogate model instead of the black-box model. Typically, the linear regression and the decision tree are usually used surrogate models because of their white-box nature. On the one hand, although the linear regression model is criticised for its disadvantages in prediction accuracy especially for non-linear problems, it can be well interpreted for every feature by quantifying both the direction and the significance of a feature’s effect. On the other hand, comparing to the linear models, tree-based surrogates inherently suit the non-linear and non-monotonic problems better, thus less loss of model performance. By performing “if-then” rules to make predictions, the tree-based surrogate model aligns with the decision process of humans, and thus it can be easily explained and understood.
    
    In this work, a Ridge regression model is used as the linear surrogate model to investigate mainly the direction of the features’ contribution in average meaning, and thus the observed coefficients are deduced equivalent to feature effects in the original black-box model. Note that the Ridge regression is an improved linear regression model with a L2 regularisation term in its loss function, which can mitigate the overfitting problem by limiting the weights of the linear regression model. Furthermore, a decision tree is used as the tree-based surrogate model to evaluate the importance of features and visualise the tree’s structure to understand how the surrogate model generates predictions step by step according to “if-then” rules. With the support of the visualised tree, it would be possible to track back the decision process for interested predictions originally made by the black-box model, and thus approximate to its internal logic.
    
    \paragraph{Partial Dependency Plots (PDPs)}
    To globally investigate the effect of interested features that influences the prediction, PDP, as discussed in the previous chapter, is a good tool to visualise it. By substituting the studied feature’s value into a range of values, changes in model prediction may imply the potential feature effect. In this paper, PDP is applied to investigate the relations between the predicted value and the feature’s value as a supporting explainer to the linear surrogate model, because the coefficients of the linear surrogate model only represent the averaged effect of a certain feature, but this kind of relationship may not be constant or even not monotonic. The advantage of applying PDP to analyse the feature effects is its intuitive and model-agnostic property, while its computation complexity for joint feature effect analysis is relatively high.
    
    \paragraph{Frequency Components Analysis}
    It is important to figure out how each of the introduced features contributes to the model performance, which also improves the model’s interpretability. Indeed, for the tree-based model, contributions of features can be quantified by checking feature importance based on criteria like information gain, Gini index, to name but a few, but we further argue that, for the time-series analysis and prediction task, the feature contribution also can be discussed not only in the time domain but also the frequency domain. By viewing the feature’s effects using spectrum analysis, the time-series model can be explained in the frequency domain. For example, if there is any feature mainly contributing to the high-frequency components, which leads to fluctuations, and any feature contributing to the DC component which determines its magnitude or trend.
    
    Therefore, inspired by the permutation feature importance measurement, this paper proposes a global interpretation technique called \textbf{Permutation Feature-based Frequency Response Analysis (PF-FRA)} to investigate feature's contribution in frequency domain. The proposed approach follows the algorithm as below:
    
    \begin{breakablealgorithm}
    	\label{algorithm}
    	\caption{Permutation Feature-based Frequency Response Analysis (PF-FRA)}
    	\begin{flushleft}
    	\textbf{Input:} Dataset with $N$ features $X = \{ {x_i};i = 1,2,3,...,N\}$\\
    	\qquad \quad Interested feature $m$ \\
    	\qquad \quad Time-series model $f(x,t;\gamma )$ \\
    	\textbf{Output:} Spectrum pair of the model response with and without the interested-feature permutation 
    	\end{flushleft}
    	\hrule
    	\begin{algorithmic}[1]
    		\State \label{algorithm-1} 
    		Train the time-series model $f(x,t;\gamma )$ by the dataset with all features $X$.
    		\State \label{algorithm-2} 
    		Based on the dataset $X$, generate a interested-feature-permutated dataset ${X_{\{ i\} /m}}$ by substituting the interested feature $m$ with its mean value.
    		\State \label{algorithm-3} 
    		Generate prediction series ${{\hat y}_{\{ i\} /m}}$ on the interested-feature-permutated dataset ${X_{\{ i\} /m}}$.
    		\State \label{algorithm-4} 
    		Compute the spectrum of ${{\hat y}_{\{ i\} /m}}$, using Fourier Transformation expressed by Eq.\ref{FourierTrans}:
    		
    		\begin{equation}
                \label{FourierTrans}
                F(w) = \int_{ - \infty }^{ + \infty } {f(t){e^{ - jwt}}dt} 
            \end{equation}
    		\State \label{algorithm-5}
    		Repeat step 2 to 4 by substituting other features with their mean values as interested-feature-remained dataset ${X_{\{ i\} /\{i-m\}}}$ to compute the spectrum of ${{\hat y}_{\{ i\} /\{i-m\}}}$.
    		\State \label{algorithm-6} 
    		Compare the spectrum pair of the model response for the two modified datasets in frequency domain.
    	\end{algorithmic}
    \end{breakablealgorithm}
    
    Since the aforementioned investigation has revealed that the RT has strong autocorrelations, it is deduced that the predicted RT should depend significantly on its previous values, and thus it is worthy to analyse the frequency contributions of the historical RT feature against the other features. it is expected to find that the historical RT may contribute more to the low-frequency components and other features contribute more to the high-frequency components that lead to the sudden changes of RT. In this way, we are supposed to approach the mechanism of how the critical historical RT feature boosts the model so significantly.
    
    \paragraph{Local Interpretable Model-Agnostic Explanations (LIME)}
    LIME\citep{ribeiro2016should} is a model-agnostic approach attempting to approach the complex model’s local behaviour through fitting a simple but explainable model like linear regression and decision tree with a set of perturbating samples around the sample that needs explanation. In this way, because we can easily understand the simple local model to generate a reason code, the prediction value given by the original complex model can be explained with the generated LIME explanations.
    
    \paragraph{Shapley Additive Explanations (SHAP)}
    SHAP\citep{lundberg2017unified} is also a model-agnostic local IML approach inspired by Game Theory\citep{myerson2013game}. In this paper, a linear SHAP is applied for our model to quantify the feature importance of a given observation and its prediction by using the additive feature attribution method which designs a linear function of binary variables approximate to the black-box model. Note that the weights of the linear function are computed based on the comparison between two models’ outputs, where one of the models is trained with all the designated features while another one is trained with the dataset withholding the evaluated feature. In this way, these weights of the linear function represent the feature importance which can be used to evaluate how much contribution of each feature is made to generate a prediction.

\section{Experimental Results Analysis and Discussion}

    \subsection{Model Evaluation and Optimisation}
    
    \paragraph{Model Evaluation and Selection}
    To fairly evaluate the applied XGBM regressor, the dataset is split into three orthogonal sets which are independent of each other as mentioned in the previous section, that is, the training set covering a 33-month period is used to train the model’s parameters, the validation set covering a 3-month period is used for hyperparameter tuning and model selection and the test set covering a 4-month period is used for model’s performance reporting. To evaluate the regression model comprehensively, we choose 4 typical metrics: Mean Squared Error (MSE), Mean Absolute Error (MAE), Mean Absolute Percentage Error (MAPE) and R-square score. 
    
    
    
    
    
    On the other hand, tuning the XGBM regressor's hyperparameters is a repetitive but critical work, which impacts the model’s final performance. It is almost impossible to find the optimum hyperparameter combination from the infinite possibilities, but the grid search method provides the way to approach it. According to certain evaluating metrics, the grid search method exhaustively tries combinations of the designated value set for the hyperparameters and then select the one with the best score on the validation set. In this work, we mainly focus on several critical hyperparameters for XGBM. Specifically, we comprehensively test the maximum depth from 5 to 15, the number of trees from 20 to 500 and minimum loss function reduction from 0.05 to 2 with reasonable intervals for the grid search to find out the best hyperparameter set.
    
    
    
\begin{figure}[htbp]
    	\centering
    	\includegraphics[width=0.8\textwidth]{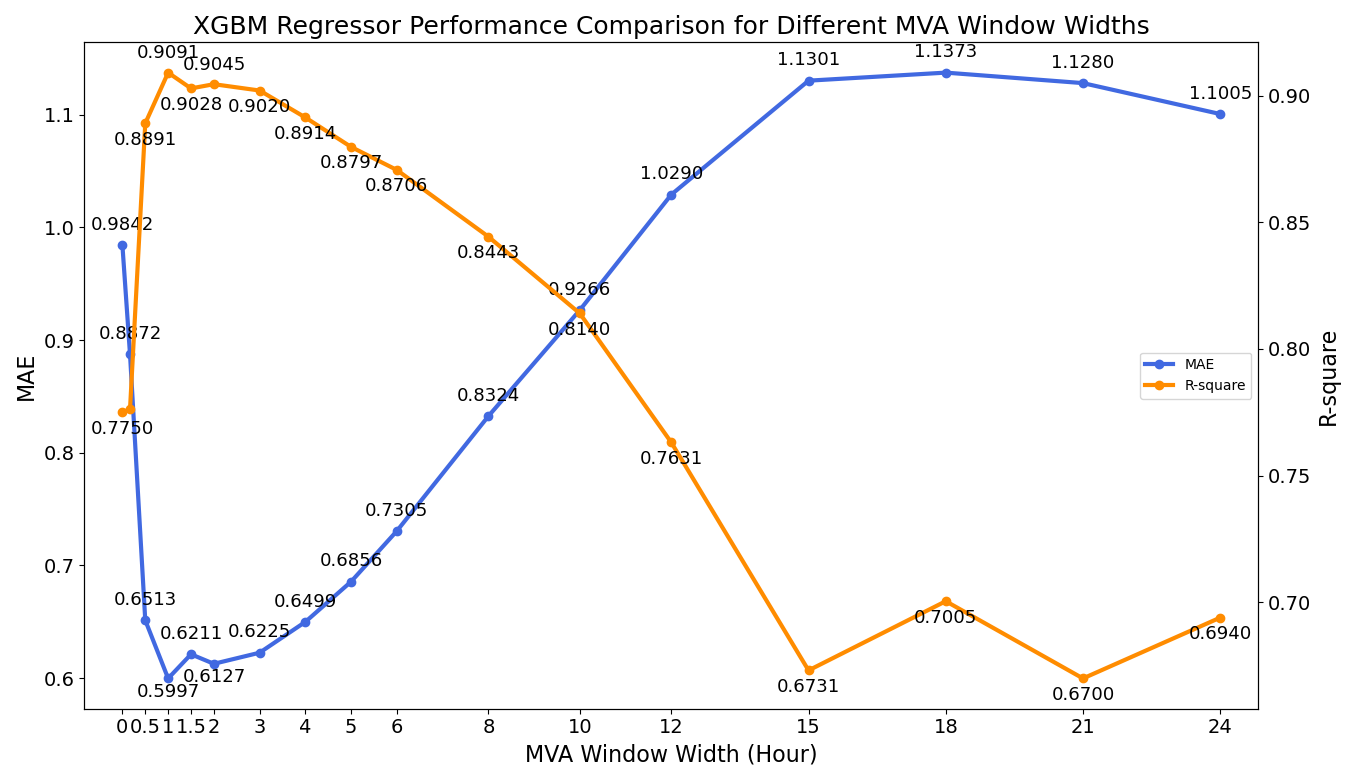}
    	\caption{Performance Comparison for Different MVA Window Width in MAE and R-square}
    	\label{MVA_window}
    \end{figure}

    \paragraph{MVA Window Width Selection}
    As discussed previously, selecting a reasonable MVA window width in the feature engineering stage is important, where a too big width can lead to information loss, but a too-small width may fail to suppress the noise and lead to model overfitting. Therefore, to select an optimum MVA window width to smoothen the temperature and humidity features with acceptable information loss and sequence delay, a performance comparison using different MVA window widths is made. Fig.\ref{MVA_window} demonstrates the performance of XGBM-based RT prediction models trained with different window widths ranging from 0 to 24 hours. As a controlling variable, the evaluations are based on models’ 8-hour ahead predictions. Noticeably, there are two special widths 0 and 10 minutes, where the width of 0 represents the model does not introduce the historical RT feature and the other continuous features are also not averaged over time, the width of 10 minutes implies only the 10-minute historical RTs is introduced as the historical feature. The comparison result indicates that the model’s performance is improved significantly after introducing the historical RT information, but at the same time, this feature becomes noisy when the MVA window width further increases. Noticeably, the model with a window width bigger than 12 hours whose performance is worse than the one without any historical RT information. Briefly, the XGBM regressor achieves the best result with the window width of 1 hour when it aims to predict 8-hour ahead RTs, which is selected as the optimum MVA window width in the later investigations.
    
    \paragraph{Predicting Time Interval Selection}
    Initially, we expect to design a fully autonomous HVAC system that can automatically control itself by accurately predicting the RT. That means the true RT should not be always accessible after system initialisation, because the RT should be treated as the output once the autonomous system starts to operate. However, because we have shown that the RT prediction model without its historical information fails in generating accurate prediction, while the accurate one significantly depends on the previous values. Therefore, it is an alternative to take a step back by considering a half-autonomous HVAC system which can acquire the true RTs periodically, for example, the RT prediction model can obtain RT's true value on a daily basis. Generally, the longer the time interval that the prediction model is allowed to access to the true RT, the closer the model approximates to the fully autonomous system, while it also implies performance drop with the increasing interval. It is critical to find out a balance for this trade-off, so Fig.\ref{Time_interval} below illustrates the MSE and MAE of the predictions given by the presented XGBM regressor over different predicting time intervals evaluated on the validation set.

   \begin{figure}[htbp]
    	\centering
    	\includegraphics[width=0.8\textwidth]{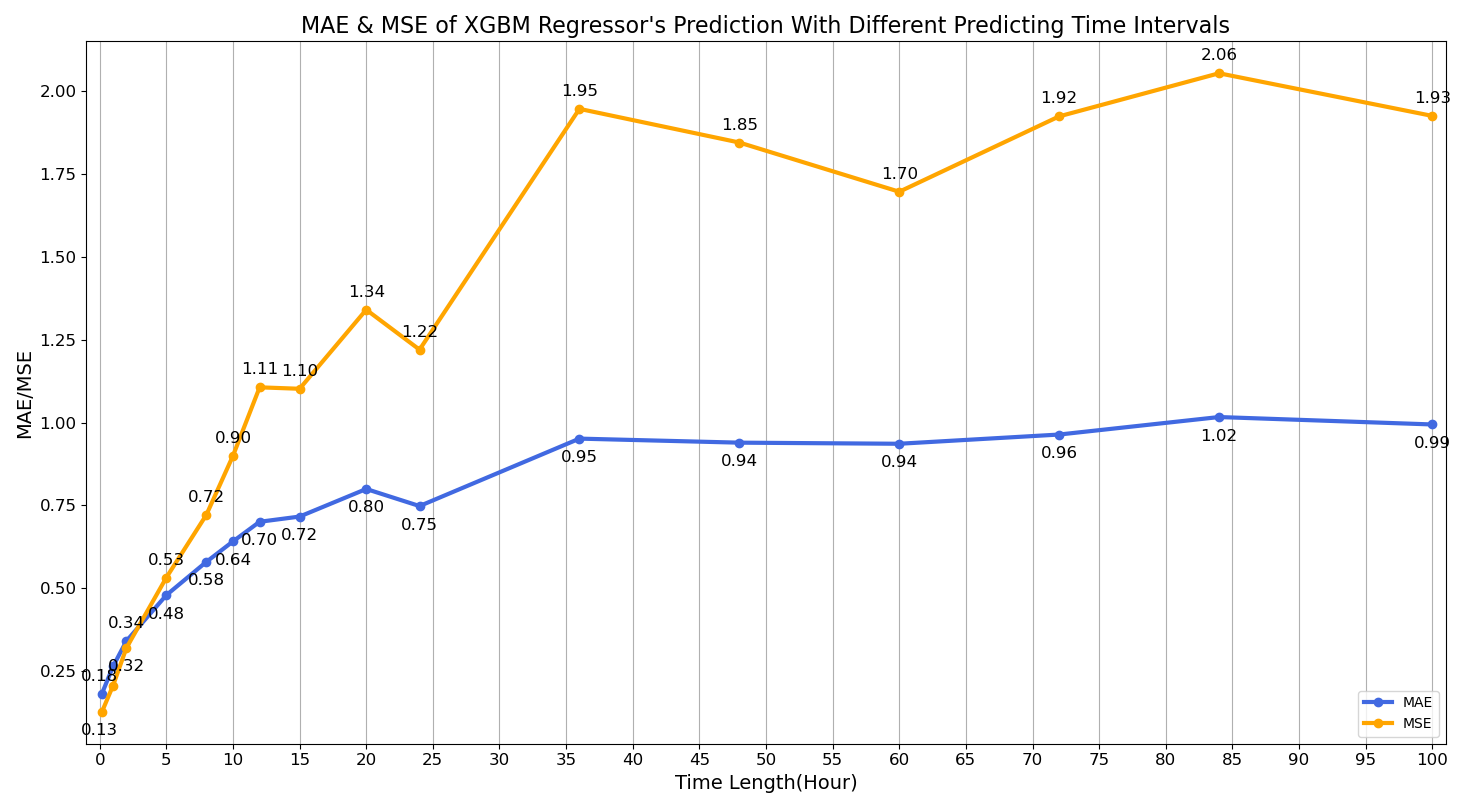}
    	\caption{MAE and MSE for RT Prediction with Different Predicting Time Intervals}
    	\label{Time_interval}
    \end{figure}   
    
    It is important to find out a balance for the trade-off between the forecasting accuracy and the degree of autonomy. We finally decide to predict 8-hour ahead RTs given a certain time observation for the true RT information, because the HVAC system needs careful control when it consumes more energy during the roughly 8-hour working hour for a day, and the performance drop is still acceptable with this predicting time interval. In this way, the HVAC system is expected to realise half-autonomous control by referring to the predicted RTs for the following 8 hours once it acquires true RT information at the start of a day.
    
    \paragraph{Feature Combinations Comparison and Selection}
    As we presented an enhanced feature engineering by not only introducing newly extracted features like historical RT and holiday information but also smoothening the temperature and humidity features with the fine-tuned MVA filter. Comparisons of using different combinations of the originally existing features and newly introduced features are made to experimentally prove the contribution of the presented feature engineering.  Furthermore, we also investigate the LSTM as the control group for comparison with the XGBM regressor to clarify the generalisation of the feature engineering’s effectiveness. Specifically, the compared features are divided into 4 groups, that is, IOTS features, IOTS features with MVA filter smoothened (IOTS-MVA), historical RT with MVA filter smoothened (MVART) and the holiday indicator. Table\ref{Performance_Comparison} summarises the performance of XGBM regressor and LSTM evaluated by the metrics of MSE, MAE, MAPE and R-square score.
    
    \begin{table}[htbp]
      \centering
      \caption{Summary of Model Performance Using Different Combinations of Features}
      \begin{tabular}{c}
        \begin{minipage}[b]{1\columnwidth}
    		\centering
    		\raisebox{-0.5\height}{\includegraphics[width=\linewidth]{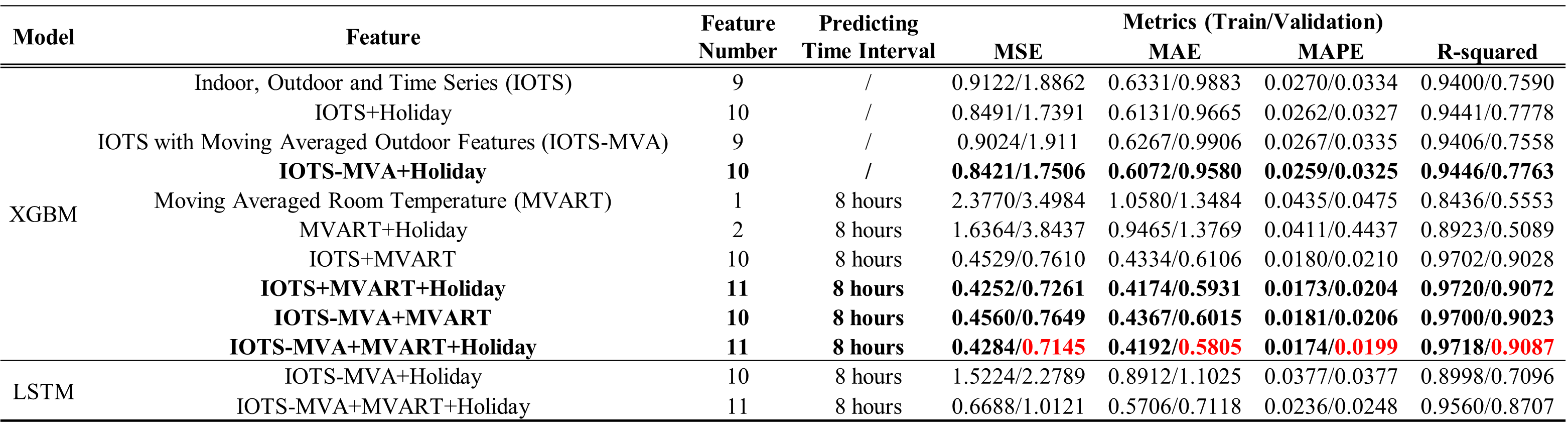}}
    	\end{minipage}
      \end{tabular}
    \label{Performance_Comparison}
    \end{table}
    
    Note that all the evaluated models demonstrated in the table are fine-tuned by using the grid search method for a fair comparison. It is found that the XGBM regressor trained with IOTS-MVA, MVART and holiday indicator has the best performance with the MAE of 0.5805°C or MAPE of 1.99\%. Importantly, comparing it with the one trained with features of IOTS-MVA and holiday indicator (the 4th row), the MVART feature contributes to the model’s performance significantly by reducing half of the error in terms of MAPE. Furthermore, the results in the 8th and 9th rows also show that the holiday indicator and MVA filter improves the model performance slightly. The comprehensive comparison experimentally proves that the enhanced feature engineering indeed improves the RT prediction accuracy to guide a more reliable half-autonomous HVAC system.
    
\begin{figure}[htbp]
    	\centering
    	\includegraphics[width=0.9\textwidth]{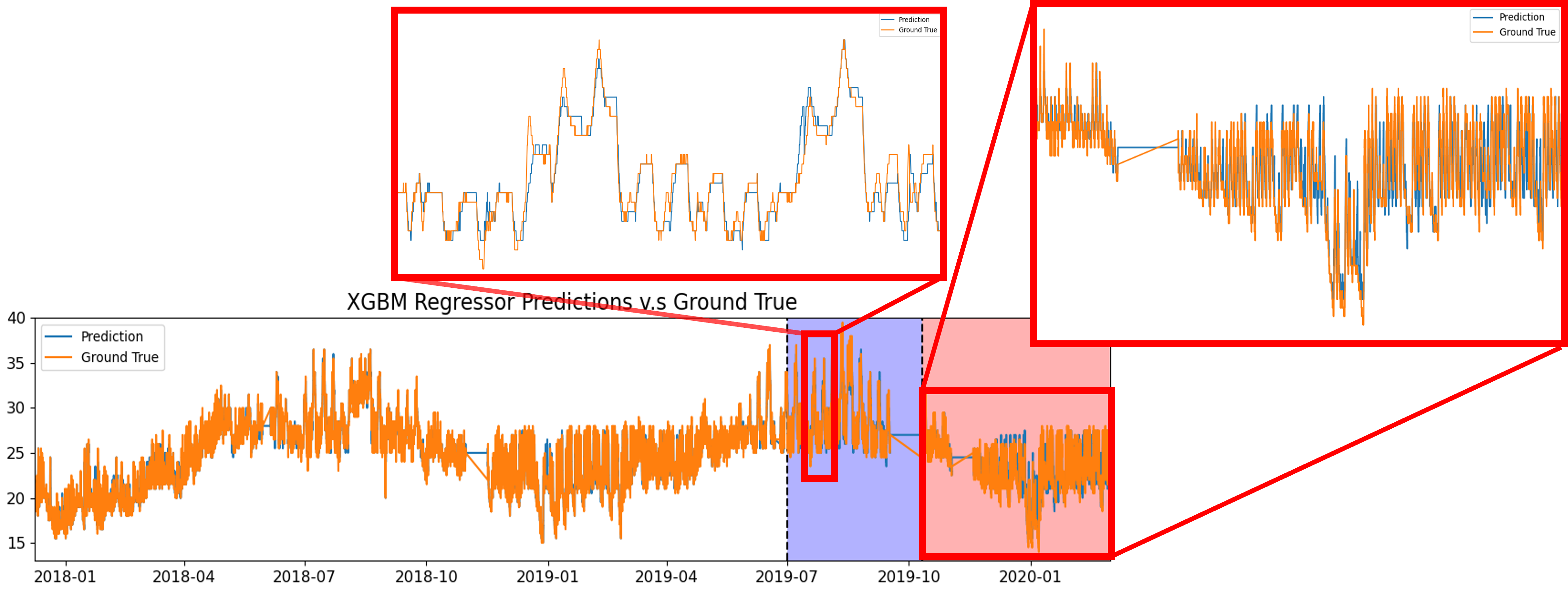}
    	\caption{Final Model’s RT Predictions versus the Ground Truth over the Whole Period}
    	\label{Final_Performance}
    \end{figure}

    \paragraph{The Final Model’s Performance}
    With the final XGBM-based RT predicting model, we then evaluate and report its performance on the test set which the model has never seen to test its generalisation. Re-training the XGBM regressor with the searched hyperparameters on the training and validation sets, it achieves 1.8073 of MSE, 0.9419 of MAE, 4.08\% of MAPE and 0.7742 of R-square score on the test set. That means, averagely, the model can predict the 8-hour ahead RTs with the error less than 1 degree, which is accurate enough result that is expected to guide the energy-efficient HVAC system control with respect to the thermal comfort. Fig.\ref{Final_Performance} compares the finalised model’s predictions with the true RT series over the whole period with a detailed plot on the validation and test sets. It is observed that the model successfully forecasts the trends as well as most of fluctuations despite failures for some extreme values. Importantly, we do not observe significant delay between the predicted and true RT series not only in the training and validation sets but also in the test set. However, comparing its performance on the test set with that on the training and validation set, overfitting still exists, which may need to be solved in the future work.
    
\begin{figure}[htbp]
        \centering
        \subfigure[Histogram Plot of the Residuals]{
        \begin{minipage}[b]{0.35\textwidth}
        \includegraphics[width=1\textwidth]{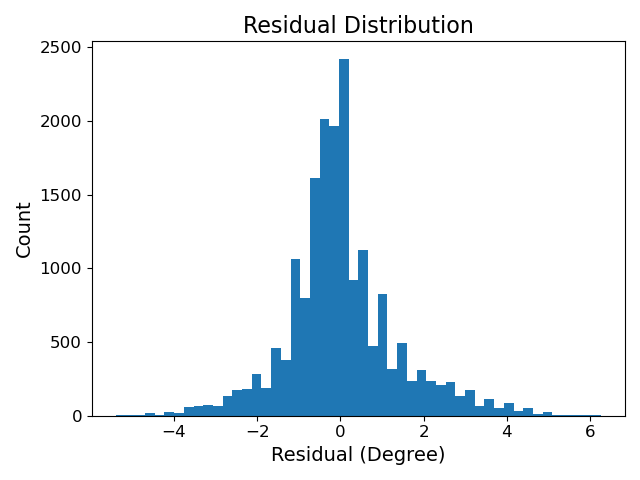} 
        \end{minipage}
        }
        \subfigure[Quantile-Quantile plot of the Residuals]{
        \begin{minipage}[b]{0.35\textwidth}
        \includegraphics[width=1\textwidth]{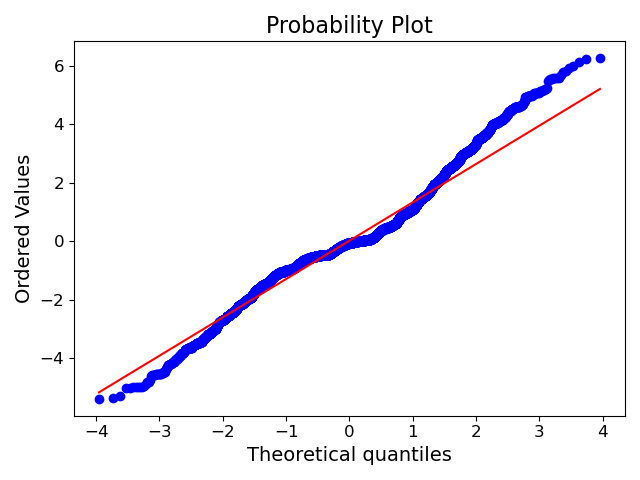} 
        \end{minipage}
        }
        \caption{Residual Analysis of the Predicted Room Temperature}
        \label{Residuals}
    \end{figure}

    To further validate the model’s bias on the test set, the distribution of residuals between the predicted and true RT is studied with the support of histogram plot and Quantile-Quantile plot (Q-Q Plot) shown in Fig.\ref{Residuals}. It can be found that the residuals are symmetrically distributed about the 0 like a Gaussian distribution, where the Q-Q plot also supports this deduction because the quantiles of the observed residuals are close to the theoretical quantiles of the Gaussian distribution in an approximately linear relationship.
    
    \subsection{Model Interpretation}
    
    Interpreting the black-box RT prediction model is so important because it not only enables the public to trust in the model itself and its predictions but also minimises the risk of applying the black-box model into such a critical system in a building through efficiently error correction according to comprehensive explanations for the model’s decision logic. In fact, we have argued the importance of increasing the model’s interpretability in previous discussions, for example, it can reduce the economic and ethical risks. Therefore, this section mainly focuses on the generated explanations by applying global IML techniques for the model’s global mechanism and local IML techniques for reason codes to understand its local behaviours.
    
    \paragraph{Model Mechanism Explanations}
    We attempt to explain the XGBM-based RT prediction model in three-fold, that is, feature importance and effects, surrogate model followed by complementary support of PDPs and frequency components analysis. Firstly, because the decision tree is used as the base estimator for XGBM model in this work, feature importance can be computed based on the information gain at the branch nodes, which can be used to quantify to what extent that the features contribute to the model. Secondly, a linear surrogate model and a decision tree-based surrogate model are trained to fit the predictions made by the XGBM model and then explained. Meanwhile, the PDPs are drawn to further observe the variation of feature effects flexibly over different values. Thirdly, we investigate the contribution of the MVART feature in frequency domain, as it contributing to the model performance significantly in the experiments. Particularly, we are interested in if the MVART feature contributes to the frequency components within a certain range.
    
\begin{figure}[htbp]
        \centering
        \subfigure[XGBM Regressor]{
        \begin{minipage}[b]{0.45\textwidth}
        \includegraphics[width=1\textwidth]{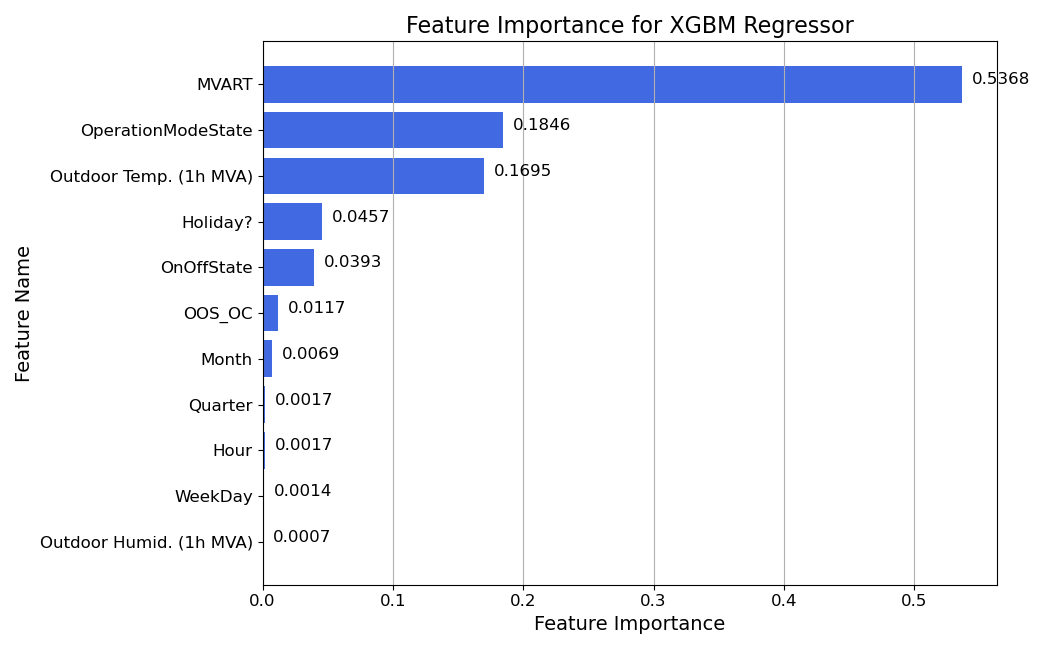} 
        \end{minipage}
        }
        \subfigure[Surrogate Decision Tree]{
        \begin{minipage}[b]{0.45\textwidth}
        \includegraphics[width=1\textwidth]{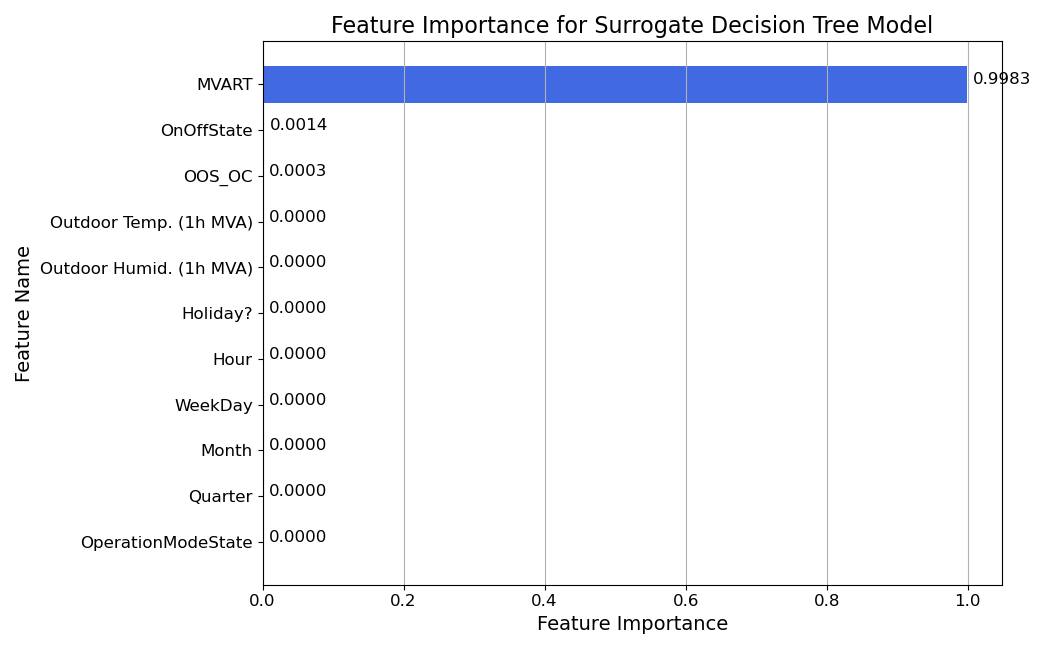} 
        \end{minipage}
        }
        \caption{ Feature Importance for the XGBM Regressor and the Surrogate Tree-based Model}
        \label{Feature_importance}
    \end{figure}

    Fig.\ref{Feature_importance} a) demonstrates the computed feature importance for the XGBM regressor, where, as expected, the historical RT information is the most important feature whose importance score reaches around 3 times the value for the second most important feature. While the HVAC system operation state and the outdoor temperature features are the 2nd and 3rd contributing features with similar effects, others like outdoor humidity and time-related features only account for very limited contribution. The results align the aforementioned analysis as well as our human intuition. For MVART, the previous stationarity analysis for RT series has presented that the value of RT at a certain time depends on its previous values, which is verified experimentally with the illustrated results. Intuitively, the RT should be influenced when the air conditioning is on, while it also should be highly positively correlated to the outdoor temperature. As an observed influential factor, the holiday effects are also verified.
    
    A surrogate decision tree model is trained with the predictions generated by the XGBM regressor to approximate the model’s decision logic. For simplification, the tree is set with a maximum depth of 6 for efficient visualisation and tracing, while this leads to most of the branch nodes choose the most important feature MVART as the split feature. Fig.\ref{Feature_importance} b) shows the feature importance for the surrogate decision tree, where it shows almost only the historical RT feature support the forecasting logic. It is deduced that this is because the decision tree tends to overfit the training data and thus only put the priority on the most informative feature. Therefore, the explanations are considered biased and unreliable. To fairly investigate the black-box XGBM model’s predicting mechanism, we use a surrogate linear regression model with L2 regularisation term (Ridge Regression), instead, to check what extent and direction that the features affect the model’s prediction averagely. The trained surrogate linear regression model can be expressed by Eq.\ref{linear_surrogate}.
    
    \begin{equation}
        \label{linear_surrogate} 
        \begin{aligned}
            & RT=-0.159\cdot OnOffState-0.03\cdot OpMode+0.037\cdot Quarter-0.011\cdot Month +0.008\cdot WeekDay \\ 
            & \qquad -0.01\cdot Hour+ 0.189\cdot OOS OC-0.052\cdot Holiday -0.004\cdot OutHumid (1hMVA) \\ 
            & \qquad +0.013\cdot OutTemp(1hMVA) + 0.944\cdot MVART +1.568
        \end{aligned}
    \end{equation}    
    
\begin{figure}[htbp]
    	\centering
    	\includegraphics[width=0.7\textwidth]{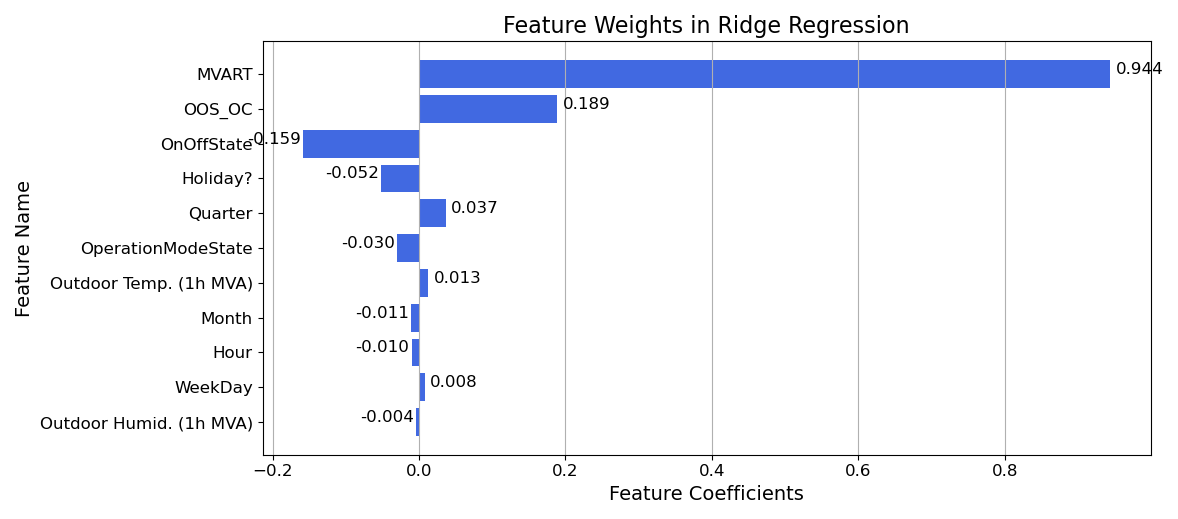}
    	\caption{Feature Coefficients for Surrogate Ridge Regression Model}
    	\label{Linear_surrogate_feature_coef}
    \end{figure}

    Or equivalently, a summary of the linear regression model’s coefficients is shown in Fig.\ref{Linear_surrogate_feature_coef}. In this case, we can conclude that, averagely, historical RT, room occupancy, quarter, outdoor temperature and the day of a week positively affect the predicted RT, but HVAC system On/Off state and operation mode, holiday indicator, month, hour and outdoor humidity negatively impact predicting values. However, these coefficients of linear regression measure the feature effects in an average way, which might be problematic if the effects are not monotonic or constant. For example, the RT should not be always negatively correlated to the month, because it is supposed to peak in the summer months but reach the bottom during the winter months. Therefore, we further draw the PDPs for each feature to find out the relationships between the model predictions and the feature value.
    
\begin{figure}[htbp]
    	\centering
    	\includegraphics[width=1\textwidth]{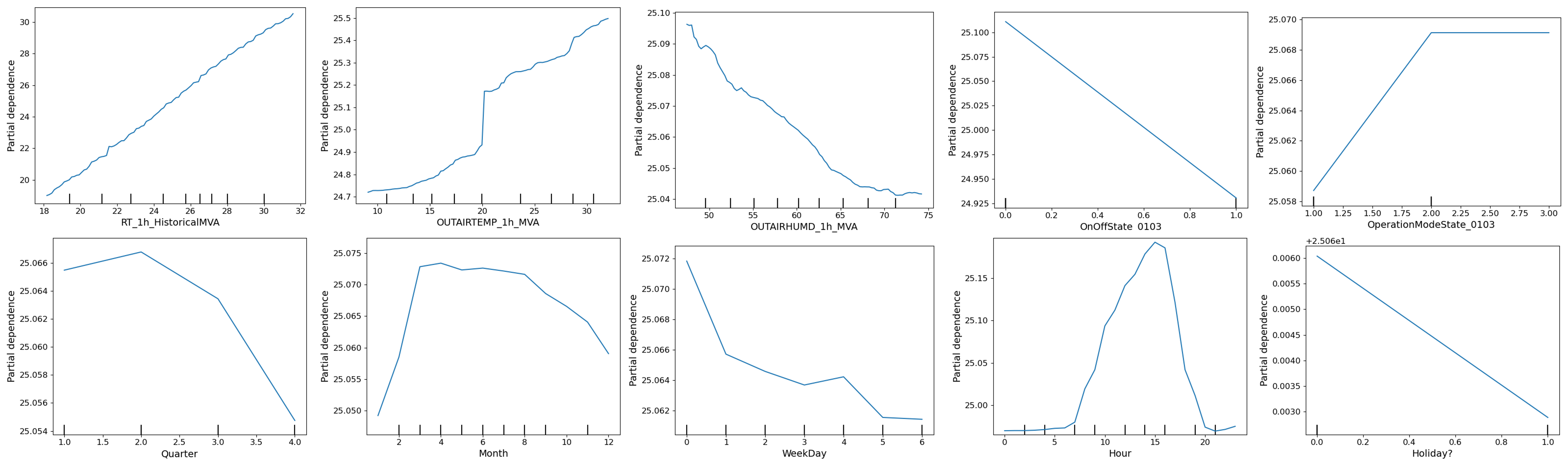}
    	\caption{PDPs for Selected Features}
    	\label{PDPs}
    \end{figure}

    Fig.\ref{PDPs} shows the PDPs for 10 out of 11 features, where the room occupancy is ignored for simplification because it is a feature extracted from the time-related feature and system On/Off state, which captures similar information. What the PDPs indicate matches our previous observations and deductions in the EDA stage, that is, the historical RT indeed has almost a linear relation with the current RT, and the outdoor temperature is also positively correlated to the RT. In contrast, the outdoor humidity has an opposite effect on the target RT value. Although other features impact the RT prediction with relatively limited effects, it can be found that the RT tends to reach the maxima during the hot time such as the second quarter (summer), months of May, June and July, the noon in a day at 11, 12 am and 1 pm, which is reasonable and intuitive for humans.
    
\begin{figure}[htbp]
    	\centering
    	\includegraphics[width=0.9\textwidth]{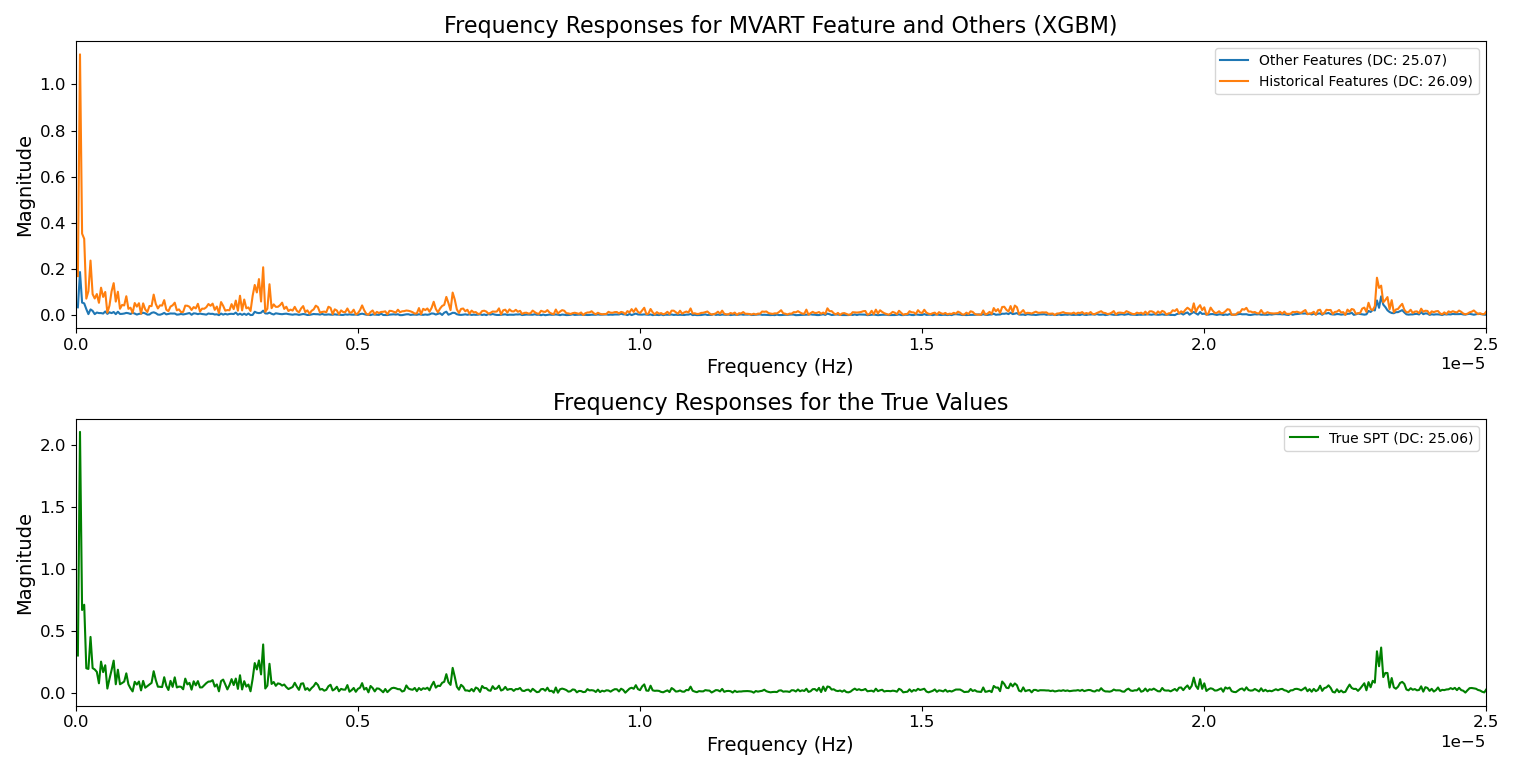}
    	\caption{Frequency Responses for the XGBM Regressor with Different Valid Features}
    	\label{PFFRA}
    \end{figure}

    Note that all the above interpretations focus on static analysis for feature effects, but it is necessary to explore more about the prediction’s characteristics in the time or frequency domain, since it is a time-series prediction task. The model’s frequency characteristics are explored by applying the proposed \textbf{PF-FRA} algorithm. As an example, since the historical RT feature is proved to be the most important feature, the MVART feature is studied to find out how the critical historical RT feature boosts the model so significantly. Fig.\ref{PFFRA} compares the XGBM regressor’s frequency responses (magnitude only) on the training and validation set with one of the MVART and IOTS features valid. The spectrum of the true RT series is also demonstrated for comparison and the DC component is not shown in the plots but indicated in the legends. It can be found that although the historical RT feature accounts for the majority of low-frequency components, both the RT and other IOTS features share a similarity for the contribution to high-frequency components. Furthermore, because the IOTS features lead to a DC component of 25.07°C closer to the original model predictions’ DC component of 25.06°C, we deduce that the IOTS features may be more informative to determine the RT’s absolute magnitude, while the MVART feature contributes to the relative variations instead.
    
    \paragraph{Case Study on Individual Predictions}
    To study the chosen RT prediction model’s local behaviour, we select several representative observations to conduct the case study. The observations are from two groups, where one is accurate group and another one is the deviated group. And the studied observations are selected accordingly, that is, the observations with a prediction error less than 0.01°C are considered as the accurate predictions but those with an error greater than 2°C are considered as the deviated predictions. In this way, we intend to figure out why the individual prediction value varies given the same target RT value. We demonstrate analysis for two observations with the same true RT value but from different groups, but the back-end model behaviour should be the same. As discussed, two local IML techniques: LIME and SHAP are applied to support each other’s explanation.
    
    \begin{figure}[htbp]
        \centering
        \subfigure[Accurate Prediction]{
        \begin{minipage}[b]{0.48\textwidth}
        \includegraphics[width=1\textwidth]{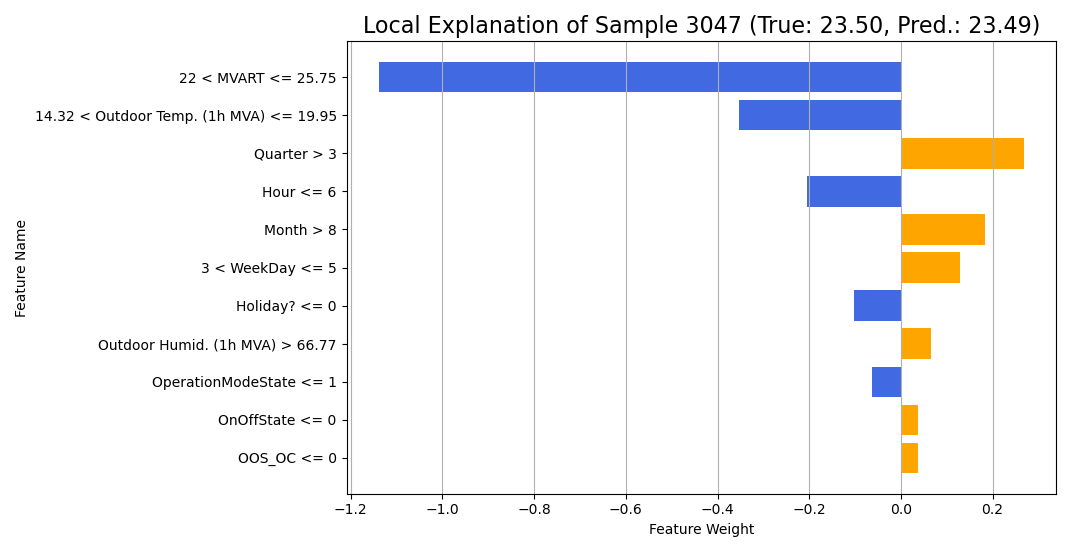} 
        \end{minipage}
        }
        \subfigure[Deviated Prediction]{
        \begin{minipage}[b]{0.48\textwidth}
        \includegraphics[width=1\textwidth]{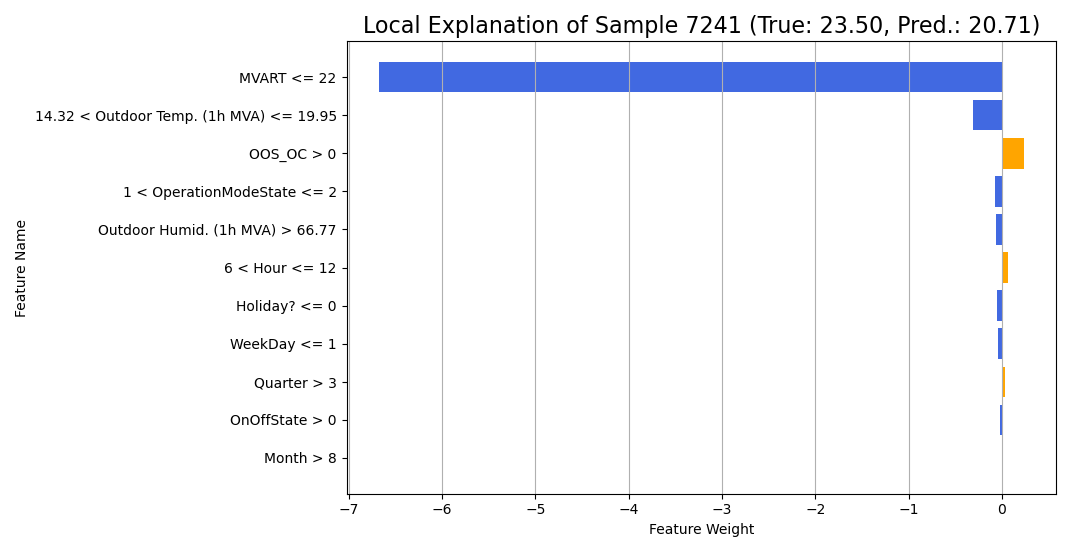} 
        \end{minipage}
        }
        \caption{LIME Explanations for Studied Observations}
        \label{LIME_res}
    \end{figure}    
    
\begin{figure}[htbp]
        \centering
        \subfigure[Accurate Prediction]{
        \begin{minipage}[b]{0.9\textwidth}
        \includegraphics[width=1\textwidth]{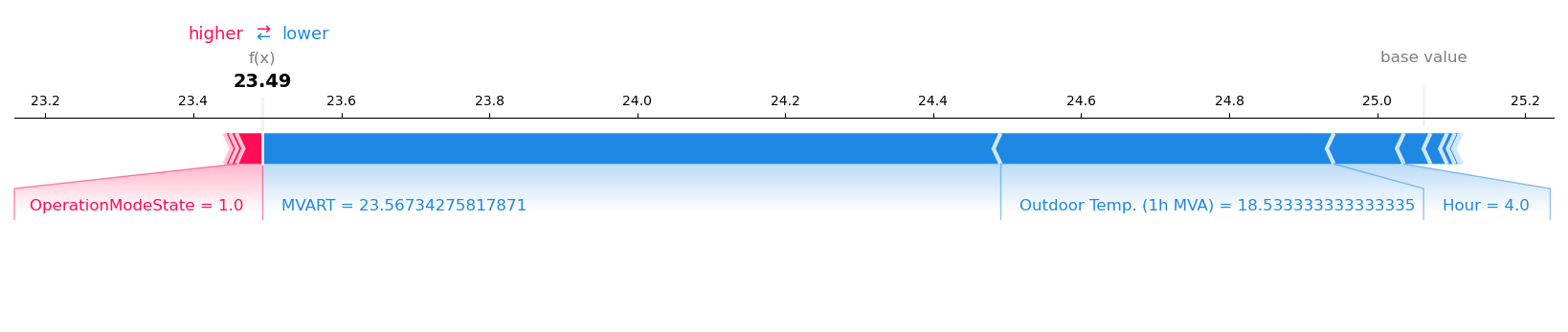}
        \end{minipage}
        }
        \subfigure[Deviated Prediction]{
        \begin{minipage}[b]{0.9\textwidth}
        \includegraphics[width=1\textwidth]{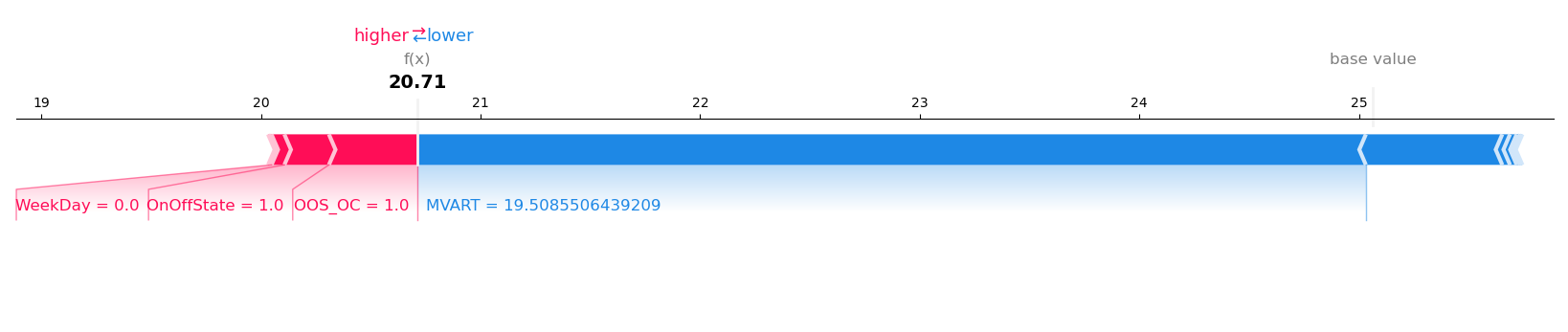} 
        \end{minipage}
        }
        \caption{SHAP Explanations for Studied Observations}
        \label{SHAP_res}
    \end{figure}

    The two studied observations are all from the test set with the same true RT value of 23.5°C, while the accurate prediction is 23.49°C and the deviated prediction is 20.71°C. Fig.\ref{LIME_res} shows the explanations given by LIME, where a) is for the accurate one and b) is for the deviated one. By comparison, it is easy to find that the main factor leading to the different predicted value is because the MVART feature falls in different value range even though most of other features have the same value. The local explanation may alarm us that the XGBM-based RT prediction model is likely to fail after a sudden change of RT values. To further validate this conclusion, we apply the SHAP to visualise feature effects for the two individual observations. Fig.\ref{SHAP_res} illustrates the force plots with the feature Shapley values. It can be observed that the for the two RTs smaller than the base value (the sample mean 25.06°C), MVART feature indeed negatively enforces the predicted value lower significantly, while it over-enforces the deviated prediction to a further lower RT because of the lower value of MVART in this case. Compared to the MVART feature, other features’ effects are not so significant as what we have argued in previous analysis. This result aligns the explanations given by LIME, clarifying that the historical RT information is so deterministic for the model’s prediction, and thus we suggest that the RT prediction model should be carefully supervised when the RT has sudden and violent fluctuations.
    
\section{Conclusion}
In conclusion, this work proposes an interpretable machine learning model to predict room temperature for non-domestic buildings with an installed HVAC system, which aims to inform a half-autonomous HVAC system controlling strategy to optimise the energy efficiency and remain thermal comfort for users. The proposed XGBM-based model is able to predict 8-hour ahead RTs accurately and in real-time with the observations from the past 1 hour. With the support of both existing and our proposed IML techniques, the black-box model is well explained to reduce the economic and ethical risk of applying an unknown machine learning model and also increase the trust of non-expert users with a better model transparency.

Based on comprehensive EDA for characteristics of the features and target variable and their joint relations in the given dataset, we identify a series of clues that may inspire the following improvement for feature engineering. Accordingly, an enhanced feature engineering is presented and experimentally proved to be effective for the afterwards RT prediction modelling. For instance, it is verified that introducing the historical RT feature contributes to the model performance significantly by not only comparing the model performance with and without this feature but also explanations generated by a set of interpretation techniques. Furthermore, the MVA filter is also efficient to denoise the original violently fluctuating continuous features thus mitigates the overfitting problem. The newly introduced holiday indicator is believed to accurately capture the holiday effects for the target RT variable.

In the modelling phase, we mainly investigate the possibility to accurately predict the room temperature, and thus it is possible to indirectly inform the HVAC system control strategy to realise a half-autonomous system. The fine-tuned XGBM model achieves its best performance in predicting 8-hour ahead RTs based on the presented enhanced feature engineering with the 0.9263°C of Mean Absolute Error (MAE) or 4.01\% of Mean Absolute Percentage Error (MAPE). Then, the finalised RT prediction model is then explained from both global and local perspectives to increase its transparency and trust. The global interpretations figure out that the historical RT with MVA is the most important and deterministic feature that contributes to the model performance as expected, while other features like system operation mode, outdoor temperature and room occupancy also affect the model’s decision logic. Noticeably, to explain the feature contribution in frequency domain, we propose the \textbf{PF-FRA} to study the spectrum characteristics of the predicted series. By applying the \textbf{PF-FRA} to interprete the XGBM regressor which aims to predict the RTs, it is found that the historical RT feature contributes to the majority of low-frequency components, while the other features are more informative to determine the RT’s absolute magnitude (the DC component). From the local perspective, we mainly demonstrate explanations of two comparable individual predictions (deviated and accurate, respectively) to check the model’s local behaviour. Based on the understanding of the model’s local behaviour, we conclude that the RT prediction model should be carefully supervised when the RT has sudden and violent fluctuations.

However, although work done so far in feature engineering, modelling and interpretation have mitigated the overfitting problem and improve the model’s reliability to design an energy-efficient HVAC system with respect to thermal comfort, we still observe overfittings and possible model failures when RT violently fluctuates. Future work may focus on further improving the model’s robustness and accuracy with the requirement of good model interpretability and trust.

\section*{Acknowledgements}
The authors would like to acknowledge and thank Angeliki Liaska whose final year project inspired the work presented in this paper. They are also very grateful to General Technology Ltd for providing the datasets, as well as to Konstantinos Karagiannis for the fruitful discussions and guidance.

\bibliographystyle{unsrtnat}
\bibliography{refs}  






\end{document}